\documentclass[msom,nonblindrev]{informs4} %dblanonrev; nonblindrev

\RequirePackage{tgtermes}
\RequirePackage{newtxtext}
\RequirePackage{newtxmath}
\RequirePackage{bm}
\RequirePackage{endnotes}

    \DoubleSpacedXI          % MSOM
    \usepackage{algorithmic}
    \usepackage[ruled,vlined,noend]{algorithm2e} 
    \SetKwInput{KwInit}{Initialize}
    \SetKwInput{KwIn}{Receive}
    \SetKwInput{KwOut}{Return}
    \SetKwFor{For}{\textbf{For} }{\textbf{do}}{}
    \SetAlFnt{\small       }
    \SetAlCapFnt{\small       }
    \SetAlCapNameFnt{\small       }
    \algsetup{linenosize=\small       }
    \usepackage{tikz}
    
		\usepackage[colorlinks=true,bookmarks=false,urlcolor=blue, citecolor=blue,linkcolor=blue,bookmarksopen=false,draft=false]{hyperref}

    \usepackage{natbib}
    \bibpunct[, ]{(}{)}{,}{a}{}{,}%
    \def\bibfont{\small}%
    \EquationsNumberedThrough    % Default: (1), (2), ...
    \TheoremsNumberedThrough     % Preferred (Theorem 1, Lemma 1, Theorem 2)
    \ECRepeatTheorems  %  

    \MANUSCRIPTNO{}
\makeatletter
\newcommand{\raisemath}[1]{\mathpalette{\raisem@th{#1}}}
\newcommand{\raisem@th}[3]{\raisebox{#1}{$#2#3$}}
\makeatother
\usepackage{scalerel,mathtools}

\newcommand{\distD}{\ensuremath{{\mathrm{D}}}} 
\newcommand{\pRV}{\ensuremath{\boldsymbol{p}}}
\newcommand{\dRV}{\ensuremath{\boldsymbol{d}}}

\newcommand{\etaVaR}{\ensuremath{\eta^{\scaleto{\mathrm{VaR}}{4pt}}_\alpha}}

\newcommand{\indexSet}[1]{\llbracket #1\rrbracket}
\newcommand{\expt}{\mathbb{E}}
\newcommand{\prob}{\mathbb{P}}
\newcommand{\indicator}{\mathbb{I}}

\newcommand{\D}{\mathcal{D}}
\newcommand{\A}{\mathcal{A}}

\newcommand{\Q}{\mathcal{Q}}
\newcommand{\I}{\mathcal{I}}
\newcommand{\R}{\mathcal{R}}
\newcommand{\pSet}{\ensuremath{\mathcal{P}}}
\newcommand{\PIP}{\ensuremath{\mathcal{P}^{\raisebox{1pt}{\scaleto{\mathrm{PI}}{4pt}}}}}

\newcommand{\thetaFTL}[1]{\tilde\theta^{#1}}
\newcommand{\piFTL}{\pi^{\scaleto{\mathrm{FTL}}{4pt}}}
\newcommand{\pFTL}{\tilde p}
\newcommand{\dFTL}{d^{\scaleto{\mathrm{FTL}}{4pt}}}

\newcommand{\piARL}{\pi^{\scaleto{\mathrm{ARL}}{4pt}}}
\newcommand{\piCI}{\pi^{\scaleto{\mathrm{CI}}{4.5pt}}}

\newcommand{\piNRM}{\pi^{\scaleto{\mathrm{NRM}}{4pt}}}
\newcommand{\pNRM}{p^{\scaleto{\mathrm{NRM}}{4pt}}}
\newcommand{\pCI}{p^{\scaleto{\mathrm{CI}}{4.5pt}}}

\newcommand{\CIR}{\text{CI-R}}

\newcommand{\FTLO}{\text{FTL-O}}
\newcommand{\ARLO}{\text{ARL-O}}
\newcommand{\CIO}{\text{CI-O}}
\newcommand{\NRM}{\text{NRM-O}}

\begin{document}

\ARTICLEAUTHORS{%
\AUTHOR{Parshan Pakiman}
    \AFF{University at Buffalo School of Management, State University of New York, Buffalo, NY, 14260, USA,
    \EMAIL{parshanp@buffalo.edu}}
\AUTHOR{Boxiao Chen, Selvaprabu Nadarajah}
    \AFF{College of Business Administration, University of Illinois, Chicago, IL, 60607, USA,
    \EMAIL{bbchen@uic.edu}, \EMAIL{selva@uic.edu}}
\AUTHOR{Stefanus Jasin}
    \AFF{Stephen M. Ross School of Business, University of Michigan, Ann Arbor, MI, 48109, USA,
    \EMAIL{sjasin@umich.edu}}
} % end of the block

\RUNAUTHOR{Pakiman et al.}

\TITLE{Adaptive Risk Mitigation in Demand Learning}
\RUNTITLE{Adaptive Risk Mitigation in Demand Learning}

\ABSTRACT{%
    \textbf{Problem definition.} %
        We study dynamic pricing of a product with an unknown demand distribution over a finite horizon. Departing from the standard no‐regret learning environment in which prices can be adjusted at any time, we restrict price changes to predetermined points in time to reflect common retail practice. This constraint, coupled with demand model ambiguity and an unknown customer arrival pattern, imposes a high risk of revenue loss, as a price based on a misestimated demand model may be applied to many arriving customers before it can be revised.
    \textbf{Methodology/results.} %
        We develop an adaptive risk learning (ARL) framework that embeds a data-driven ambiguity set (DAS) to quantify demand model ambiguity by adapting to the unknown customer arrival pattern. Initially, when the number of arrivals is low, the DAS includes a broad set of plausible demand models, reflecting high ambiguity and a greater risk of revenue loss. As new data is collected through pricing, the DAS progressively shrinks, capturing the reduction in model ambiguity and associated risk. We establish the probabilistic convergence of the DAS to the true demand model and derive a regret bound for the ARL policy that explicitly links revenue loss to the amount of data required for the DAS to identify the true model with high probability. The dependence of our regret bound on the customer arrival pattern is unique to our constrained dynamic pricing problem and contrasts with no-regret learning environments, where regret is constant, independent of the arrival pattern. Relaxing the constraint on infrequent price changes to the unconstrained case, we show that the ARL policy attains a known constant regret bound. Moreover, our numerical tests show that the ARL policy outperforms benchmarks that prioritize either regret or risk alone, by adaptively balancing both without knowing the arrival pattern.
    \textbf{Managerial implications.} % 
        Adapting the level of risk aversion to the sequentially arriving sales data in pricing decisions is crucial to achieve high revenues and low downside risks, especially when prices are sticky and both the customer demand distribution and their arrival pattern are unknown.}\looseness=-1
\HISTORY{\today \ (current version)}
\maketitle

        \section {Introduction} 
    
   We study a classical revenue management problem in which a seller dynamically prices a product over a finite horizon. The seller faces model ambiguity, meaning some parameters of the customer demand distribution are unknown but lie in a known finite set. The seller can learn these parameters from observed sales data to refine its pricing decisions. Dynamic pricing with demand learning is a well-established topic in the literature (e.g., \citealt{bitran2003overview, elmaghraby2003dynamic, araman2010revenue, keskin2014dynamic, den2015dynamic, chen2015recent, Feng_How_2018, shin2023dynamic}) and is widely used in practice \citep{mckinsey_dynamic_pricing_chemicals}. Most existing studies assume a large number of periods, where price adjustments are allowed in each period. This environment can be viewed as pricing at the individual customer level when the number of periods corresponds to the number of customers. 

    In this paper, we investigate a distinct environment in which price adjustments are a priori restricted to predetermined times, meaning they are not allowed at the individual customer level. This constrained pricing setting is relevant, for example, to omnichannel retailers striving to synchronize online and in-store prices to minimize customer frustration over price discrepancies (see, e.g., \citealp{McKinsey_2016} and \citealp{WE_2020}). Maintaining price consistency across channels often necessitates less frequent online price changes, ensuring a seamless shopping experience for customers. In addition, our setting arises in brick-and-mortar stores selling fast-fashion products, where selling seasons typically last two to three months. Physical stores often maintain fixed prices throughout the day to avoid intra-day updates that could frustrate customers or impose additional setup costs. Infrequent price changes distinguish our setting from the standard demand‐learning literature and require mitigating the risk of revenue loss that arises when many customers, arriving in an unknown pattern, encounter suboptimal prices optimized based on misspecified demand parameters. Such prices can yield substantially lower cumulative revenue than the optimal price, which requires knowing the true demand distribution.\looseness=-1
    
    In the standard demand learning environment with frequent price changes (i.e., the unconstrained setting), a common approach to jointly learn the true demand parameters and optimize price is the follow-the-leader learning approach (FTL; \citealp{shalev2011online}). FTL builds a maximum likelihood estimate of the true demand parameters using past sales data and then selects the price that maximizes the revenue function under this estimate. The FTL policy achieves no regret: its expected cumulative revenue loss from not knowing the true demand parameters converges to zero as the number of arrivals increases. However, in our constrained setting, the FTL policy can result in significant revenue losses due to infrequent price changes. Specifically, during periods of high model ambiguity and many arrivals, the revenue under a price based on a misestimated parameter can be substantially lower than that under the optimal price based on the true parameters. Because it is not possible to directly anticipate when model ambiguity will be high or when large arrivals will occur, the regret of the FTL policy can be substantial depending on the arrival pattern.\looseness=-1
    
\begin{figure}[t]
    	\centering
    	\caption{Illustration of customer arrival patterns (top) and evolution of the DAS over time across patterns (bottom), with darker colors showing a higher probability of a parameter being included in the DAS.\looseness=-1}\vspace{4pt}
    	\includegraphics[width=1\linewidth]{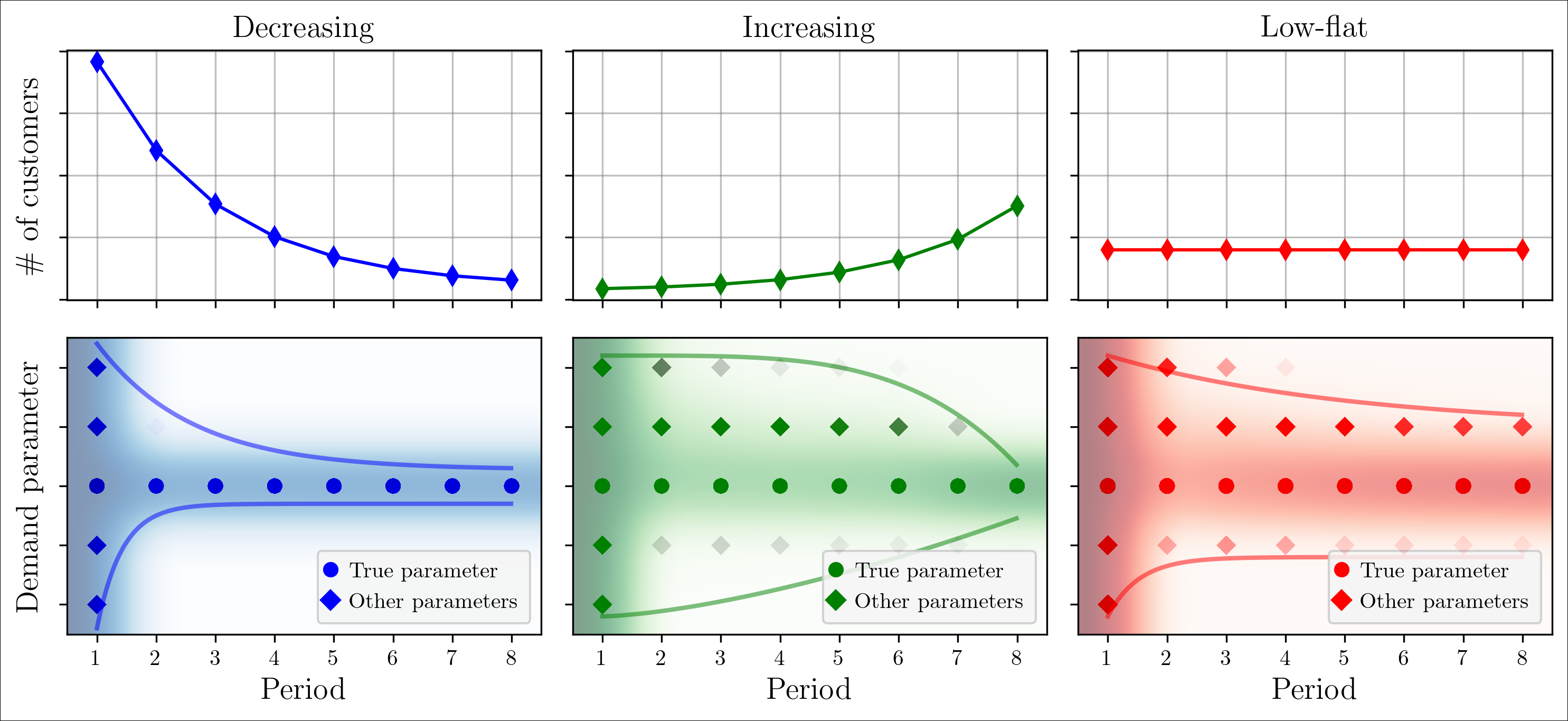}
    	\label{fig:pattern_DAS_evolution}\vspace{-30pt}
    \end{figure}

    Figure \ref{fig:pattern_DAS_evolution} presents, in its top panel, three representative customer arrival patterns: decreasing, increasing, and low-flat, which correspond to a product that becomes an early hit, a delayed hit, or a flop, respectively. In each plot, the x-axis denotes the time period, the y-axis shows the number of customer arrivals, squares mark the actual arrivals in each period, and the curve depicts the overall trend in arrivals. Intuitively, the period by which the true demand parameter is identified with high probability occurs earlier in the early-hit pattern than in the flop scenario. Once again, the arrival pattern and the time at which collected data suffice to identify the true demand parameter with high probability are unknown ex ante. Thus, a method is needed that quantifies demand model ambiguity from data arrivals and adaptively transitions from risk minimization when ambiguity is high to revenue maximization as it diminishes.

    We propose an Adaptive Risk Learning (ARL) framework that, without knowledge of the customer arrival pattern, quantifies demand model ambiguity from arriving data and dynamically sets prices to balance risk and revenue. The ARL policy selects prices that maximize a risk-adjusted revenue function defined with respect to a data-driven ambiguity set (DAS) of demand models. At each time period, the DAS consists of demand parameters that remain statistically plausible candidates for the true parameter. Early in the selling horizon, when data is limited and demand model ambiguity is high, the DAS includes many candidate parameters, including the true one. As additional data is collected through pricing, the DAS contracts to retain only those parameters that remain consistent with the observed data with high probability. This adaptive stochastic behavior of the DAS allows ARL to modulate its level of risk aversion based on the observed customer arrival pattern, without requiring any prior knowledge of that pattern. Furthermore, by extending the point-estimate approach used in FTL to a set estimation, ARL accesses a distribution over plausible revenue values rather than a single revenue value, allowing ARL to optimize prices based on different risk measures. As a result, ARL (i) mitigates the seller’s exposure to revenue loss when demand model ambiguity is high and (ii) seamlessly transitions to a regret-minimizing strategy as ambiguity declines with accumulating data.\looseness=-1
    
    The bottom panel of Figure \ref{fig:pattern_DAS_evolution} illustrates how the DAS evolves over time under each of the three customer arrival patterns depicted in the top panel. In each plot, the x-axis represents the time period, and the y-axis indexes five candidate parameters. Shaded regions indicate the probability of a parameter being included in the DAS, with darker shading corresponding to higher inclusion probability. Markers identify individual demand parameters: circles denote the true parameter, while diamonds represent other candidate parameters. The curves trace the evolution of the DAS, which shrinks over time in response to observed data, with the rate and shape of shrinkage varying by arrival pattern. As shown, the true parameter remains included in the DAS throughout the horizon. These plots highlight that under the early-hit pattern, where customer arrivals are concentrated early, the DAS converges more quickly to the true parameter. In contrast, for the delayed-hit and flop patterns, identification of the true parameter occurs later, showing slower data accumulation. These visualizations highlight that the DAS adapts to unknown customer arrival patterns to monitor model ambiguity.\looseness=-1

\subsection {Contributions}
    The contributions of this paper are highlighted below:

    \begin{enumerate}
        \item We extend FTL, which is suitable for unconstrained settings with frequent price changes, to ARL for environments with limited price adjustments by replacing point estimates of the true demand parameter with set estimates based on the DAS. This extension enables ARL to actively balance between regret and risk, unlike FTL, which focuses solely on regret. We establish the probabilistic convergence of the DAS: it initially includes all candidate parameters and gradually shrinks to a singleton containing the true parameter with high probability as more data becomes available. Building on this result, we derive a regret bound for the ARL policy that captures the effect of restricting price adjustments to predetermined times. Our regret bound explicitly depends on the customer arrival pattern and may not vanish as the number of arrivals grows, unlike the arrival-agnostic constant bound for FTL in unconstrained pricing settings (e.g., \citealt{CSW2017}). We also show that ARL recovers a similar constant regret if frequent price changes are permitted. Importantly, ARL delivers these theoretical results without requiring the separability assumption considered by \cite{CSW2017}.\looseness=-1

        \item We develop further analytical results to uncover the value of ARL’s adaptation mechanism. Specifically, we compare the ARL policy against a non-adaptive risk-mitigating (NRM) policy, which optimizes the risk-adjusted revenue over all candidate parameters, and with an FTL policy. Compared to NRM, ARL’s revenue assessments are closer to the true revenue with high probability. Also, NRM achieves a regret bound that is agnostic to the customer arrival pattern, while ARL achieves a stronger regret bound that vanishes under favorable patterns. Compared to FTL, ARL’s worst-case revenue assessment is closer to the true revenue. Lastly, FTL attains vanishing regret under favorable arrival patterns only under a restrictive separability condition, whereas ARL achieves this without requiring such an assumption. Thus, ARL mitigates conservative revenue assessments and delivers strong regret bounds by adapting its level of risk aversion to the arriving data, unlike NRM, and by extending FTL’s point estimation to set estimation, outperforming both benchmarks.\looseness=-1

        \item We provide a computational study that demonstrates the benefits of the ARL policy across a wide range of test instances. Our results show that ARL consistently achieves higher average revenues and lower value-at-risk compared to both the NRM and FTL policies. These improvements hold across different customer arrival patterns and designs of the candidate demand parameters. The results reveal interesting insights. In instances where the customer arrival pattern provides the seller with an opportunity to learn the true demand parameter (e.g., top-left subplot of Figure \ref{fig:pattern_DAS_evolution}), ARL achieves strong average revenue, outperforming the learning-focused FTL policy while also delivering substantially lower downside risk. In contrast, when the arrival pattern leaves little room for learning (e.g., top-right subplot of Figure \ref{fig:pattern_DAS_evolution}), the ARL matches the strong downside-risk performance of the risk-focused NRM policy and significantly outperforms FTL under this metric. These findings highlight the significant value of adaptively adjusting the level of risk aversion to the unknown customer arrival pattern in order to balance regret and risk, a capability unique to the ARL policy.\looseness=-1
    \end{enumerate}

\subsection{Novelty and Related Literature} \label{literature}

    Dynamic pricing with online demand learning is well studied. Among learning methods, first-moment matching, which quantifies the gap between empirical and model-based means, is a widely adopted approach (e.g., \citealt{BZ2009, BZ2012, WDY2014, BZ2015, CSW2017, CCA2019, CJD2018, LJS2019, CCS2019,den2019dynamic}). We leverage this point-estimation device to define our data-driven ambiguity set (DAS). The FTL policy and its extensions form an important class of online learning algorithms \citep{blum1998line,kalai2005efficient}, with first-moment matching commonly used for demand learning (e.g., \citealp{BZ2009,BZ2012,CSW2017}).  A common assumption in this literature is that prices can be adjusted at any time, thereby justifying a main focus on revenue maximization. In contrast, we study a new setting in which price adjustments are restricted to predetermined time points, a constraint informed by practice. This restriction introduces the challenge of jointly optimizing revenue and managing downside risk, which our proposed ARL framework is designed to address.\looseness=-1
    
    A few studies have examined dynamic pricing with demand learning under limited price changes \citep{broder2011online, CSW2017, chen2020data, perakis2024dynamic}. These works extend no-regret learning frameworks by penalizing frequent price experimentation, effectively placing a cap on the number of allowable price adjustments. However, they permit the timing of price changes to be chosen adaptively by the algorithm, allowing for flexible exploration throughout the selling horizon. In contrast, we consider a setting in which price changes are restricted not only in number but also in timing, limited to a small set of pre-specified periods that are not selected by the algorithm (i.e., seller). This hard constraint arises in practice (e.g., omnichannel and brick-and-mortar settings) and limits the applicability of existing policies and their regret guarantees. We extend this line of research by introducing a constrained dynamic pricing problem that nests the known unconstrained model as a special case, and by developing a tailored policy with a regret bound that reflects the impact of the hard constraint through its dependence on the customer arrival pattern.\looseness=-1

    Few papers have explored alternative notions of risk in the context of dynamic pricing with demand learning. For example, \cite{den2019dynamic} considers a form of risk arising from strategic customer behavior, where demand decreases or increases depending on whether the selling price exceeds or falls below a time-varying reference price that is unknown to the seller, in addition to the unknown demand model. \cite{chen2023robust} examines a different notion of risk associated with outlier demand realizations that may follow an arbitrary, possibly adversarial, model rather than the typical stochastic demand model. Both of these studies differ substantially from our work, which focuses on the risk stemming from restrictions that limit price changes to a pre-specified set of time points.\looseness=-1

    A complementary literature studies dynamic pricing with demand learning where the true demand distribution is non-stationary (e.g., \citealt{besbes2011minimax,besbes2015non,keskin2017chasing}). Related work in inventory control analyzes learning with shifting or cyclic demand distributions \citep{chen2021data,gong2024bandits,keskin2025nonstationary}. We instead focus on stationary demand and abstract from inventory decisions to isolate the effect of infrequent price changes on regret, demand learning, and policy design.\looseness=-1

    Moreover, our paper contributes to the growing interest in algorithms that incorporate ``adaptation’’ to reduce reliance on heuristic tuning. For example, \cite{shah2020reinforcement} develop an algorithm that dynamically balances exploration, approximation, and supervision in a reinforcement learning setting, and \cite{pakiman2020self} propose an approximate dynamic programming method that adapts its approximation architecture and hyperparameters to a given Markov decision process, eliminating manual design. We introduce ARL, an adaptive risk-mitigation mechanism that automatically balances regret and risk without heuristic alignment of these objectives. Without ARL, one can tune this trade-off case by case, but our numerical experiments show that such heuristics do not generalize across problem instances.
    % Moreover, our paper contributes to the growing interest in algorithms that incorporate ``adaptation'' to reduce reliance on heuristic tuning. For example, \cite{shah2020reinforcement} developed an algorithm that dynamically navigates exploration, approximation, and supervision in a reinforcement learning setting. \cite{pakiman2020self} proposed an approximate dynamic programming algorithm that adapts its approximation architecture and hyperparameters to a given Markov decision process, without requiring a human to design them. Our research introduces ARL, an adaptive risk mitigation mechanism that automatically balances regret and risk without heuristic alignment of these objectives. Without ARL, one can manually tune this trade-off on a case-by-case basis. However, as our numerical experiments confirm, such heuristics fail to perform well across problem instances.\looseness=-1
    
\subsection{Structure} 
    In \S\ref{sec:Problem Formulation}, we formulate the constrained dynamic pricing problem and formalize the risk of revenue loss. In \S\ref{sec:ARL-freamwork}, we introduce ARL and establish the stochastic convergence of the DAS along with a regret bound for the ARL policy. In \S\ref{sec:value-of-adaptation}, we compare ARL, NRM, and FTL policies in terms of revenue assessments and regret. In \S\ref{sec:numbers}, we present our numerical results. We conclude in \S\ref{sec:conclusion}. All proofs are provided in an online supplement.
        \section {Constraint Dynamic Pricing with Demand Learning}\label{sec:Problem Formulation}

    We consider a firm selling a product over a finite horizon of $T$ periods. At the beginning of each time $t \in\{ 1, 2, \ldots, T\}$, the firm chooses a price $p_t$ from the set of $n\ge1$ possible prices, denoted $\pSet:= \{p^1, \cdots, p^n\}$. During period $t$, a total of $N_t$ customers arrive, each encountering the price $p_t$, as this price cannot be changed once chosen at the beginning of the period. The $i$-th customer purchases a random quantity $\dRV_{ti}(p_t)$ of the product. At the end of period $t$, the firm observes the total demand $\sum_{i=1}^{N_t} d_{ti}(p_t)$ and the corresponding revenue $p_t \sum_{i=1}^{N_t} d_{ti}(p_t)$, where $d_{ti}(p_t)$ denotes a realization of $\dRV_{ti}(p_t)$. The number of customers $N_t$ becomes known only at the end of period $t$, meaning it is unknown when the price $p_t$ is set.  Define the index set $\indexSet{l}$ for any integer $l \geq 1$ as $\indexSet{l} := \{1,2,\dots,l\}$. For each time $t \in \indexSet{T}$ and price $p_t \in \pSet$, we assume that the realized demand $d_{ti}(p_t)$ for each customer $i \in \indexSet{N_t}$ is an i.i.d. sample from a fixed distribution $\distD(p_t)$. That is, customer demands are mutually independent and time-invariant. We refer to $\distD(\cdot)$ as the true demand distribution and impose the following mild assumption on its richness, which covers all light-tailed and most heavy-tailed distributions observed in practice \citep[Chapter 3]{foss2011introduction}.\looseness=-1

    \begin{assumption}\label{assump:LightTailedDemand} 
    	The true demand distribution is sub-exponential, meaning there are parameters $v,b\ge0$ such that $\expt[\exp(\lambda \distD(p))] \le \exp({\lambda^2 v^2}/{2})$ holds for every $\lambda\in(-{1}/{b},{1}/{b})$ and $p \in\pSet$.
    \end{assumption}

    The firm's objective is to find a pricing policy $\pi$, which is a collection of decision rules $\{\pi_t: t \in \indexSet{T}\}$, that maximizes expected cumulative revenue over the planning horizon. Each decision rule $\pi_t$ maps its own past pricing decisions and realized demands up to time $t-1$ to a price at time $t$, denoted by the random variable $\pRV_t^\pi \in \pSet$. In the complete information (CI) environment, where the firm knows the true demand distribution, the expected cumulative revenue of policy $\pi$ can be computed as follows:
    \[
    \CIR(\pi):= \expt^\pi \left[ \sum_{t=1}^T \pRV_t^\pi \cdot \sum_{i=1}^{N_t} \dRV_{ti}(\pRV_t^\pi) \right].
    \]
    The expectation is taken over the sequences of demand realizations and pricing decisions under $\pi$. In the CI setting, the policy that maximizes $\CIR(\pi)$ charges a fixed price in every period. We call this the CI policy and denote its price by $\pCI$, which solves the static problem $\max_{p \in \pSet}\{p \expt[\distD(p)]\}$. The CI policy results in the expected cumulative revenue $\CIR(\piCI)=\pCI \expt[\distD(\pCI)]M$, given by the product of the per-customer revenue $\pCI \expt[\distD(\pCI)]$ and the constant $M:=\sum_{t=1}^{T} N_t$, which is the total store traffic. Unfortunately, the firm faces demand model ambiguity, as the true demand distribution is unknown and should be learned from data. \looseness=-1
    
    Although the true demand distribution is unknown, it is reasonable to assume that the firm can use domain knowledge to specify a class of plausible demand candidates for $\distD(\cdot)$. We assume the true demand distribution admits the form of $\distD(\cdot):= \distD(\cdotp;\theta^0) + \boldsymbol{\zeta}$, where $\theta^0$ denotes the true demand parameter, $\distD(\cdotp;\theta^0)$ is a random variable encoding mean demand, and $\boldsymbol{\zeta}$ is a mean-zero random variable capturing demand noise. We also associate with this model the finite candidate set $\Theta := \{\theta^0,\theta^1,\dots,\theta^m\}$ containing the true demand parameter and $m\ge0$ other plausible candidates. Under this model, the demand model ambiguity faced by the firm stems from not knowing which parameter $\theta$ in the candidate set $\Theta$ is the true parameter $\theta^0$. This specification is fairly general, as $\distD(\cdot;\theta^0)$ may take any functional form and $\Theta$ may be arbitrarily large (though finite), provided Assumption \ref{asm:deman-model-set} holds. It ensures that for every two distinct $\theta,\theta'\in\Theta$, there exists a price $p\in\pSet$ such that $\mu(p;\theta)\neq \mu(p;\theta')$. Assumption \ref{asm:deman-model-set} is nonrestrictive, since if two parameters induce identical mean demands at all prices, one can be removed from $\Theta$ without loss. For each $\theta\in\Theta$ and $p\in\pSet$, define the mean demand $\mu(p;\theta):=\mathbb{E}[\distD(p;\theta)]$ and the expected revenue $r(p;\theta):=p,\mu(p;\theta)$.\looseness=-1
    \begin{assumption}\label{asm:deman-model-set}
        For every pair of distinct demand parameters $\theta,\theta'\in\Theta$, there exists a price $p\in\pSet$ such that identity $|\mu(p;\theta)- \mu(p;\theta')|>0$ holds.
    \end{assumption}
    
    In our constrained dynamic pricing problem, the firm faces a high risk of revenue loss due to the demand model ambiguity, the unknown customer arrival pattern, and the limited price adjustments. To elaborate, suppose the firm mistakenly estimates $\bar{\theta} \neq \theta^0$ as the true parameter and selects at time $t$ the price $p^{\bar{\theta}} \in \arg\max_{p \in \pSet} r(p;\bar{\theta})$ that maximizes expected revenue under this assumption. Because all $N_t$ customers arriving in period $t$ encounter the same suboptimal price $p^{\bar{\theta}}$, a substantial portion of cumulative revenue will be lost if the number of arrivals $N_t$ is large relative to total traffic $M$. Under the worst-case customer arrival pattern, when every customer arrives in period $t=1$ so that $M = \sum_{t=1}^T N_t = N_1$, the expected loss equals $M \bigl(r(p^{\bar{\theta}};\theta^0) - r(\pCI;\theta^0)\bigr)$, which scales linearly in $M$. Under a favorable pattern, where customers arrive uniformly over time, this loss may be sublinear if the firm can use early arrivals to reduce model ambiguity quickly, gain confidence that a candidate $\bar{\theta}$ matches $\theta^0$, and leverage later arrivals to maximize revenue.\looseness=-1 
    
    This arrival-dependent behavior of revenue loss in our setting is in sharp contrast to the unconstrained dynamic pricing problems studied in the literature, in which the price can be essentially updated per customer (i.e., assume $N_t=1$ in our setting), resulting in a revenue loss that is constant and independent of $M$ under any arrival pattern (e.g., see Proposition 1 in \citealt{CSW2017}). Therefore, in what follows, we propose a new joint pricing and demand learning framework that actively mitigates the firm’s risk exposure by adapting the level of risk to the a priori unknown arrival pattern $(N_1,N_2,\dots,N_T)$.
        
\section{Adaptive Risk Learning Framework}\label{sec:ARL-freamwork}

    In this section, we present the ARL algorithm, which both mitigates the risk of revenue loss and learns the true demand model using an ambiguity set of candidate models. In \S\ref{sec:risk-adjust-rev}, we show how to adjust the revenue function for risk using the ambiguity set. In \S\ref{sec:partially-inform-price}, we introduce partially informative prices that enable ARL to identify the true demand model. In \S\ref{sec:alg}, we detail the algorithm and the data-driven construction of the ambiguity set (i.e., DAS). In \S\ref{subsec:ARL-theory}, we analyze the theoretical properties of ARL, including the stochastic convergence of the DAS and the ARL policy's regret bound.\looseness=-1
    
\subsection{Adjusting Revenue for Risk}\label{sec:risk-adjust-rev}

    We utilize the ambiguity set $\A \subseteq \Theta$ to encode the demand parameters that the firm considers plausible candidates for the true parameter. Consider two extremes: $\A = \Theta$ and $\A = \{\theta^0\}$. The former reflects full ambiguity, treating every model in $\Theta$ as equally plausible for $\theta^0$, while the latter reflects no ambiguity, assuming the firm essentially knows $\theta^0$. Thus, the use of ambiguity set enables us to model a spectrum of demand model ambiguity, which is high when $\A = \Theta$ and low when $\A = \{\theta^0\}$. Any intermediate set $\A$ satisfying $\{\theta^0\} \subsetneq \A \subsetneq \Theta$ corresponds to a level of model ambiguity between these two extremes. We can use the ambiguity set $\A$ to adjust for the risk of revenue loss due to demand model ambiguity. Given a price $p$, we define the set of plausible expected revenues $\R(p;\A) := \{ r(p; \theta) : \theta \in \A \}$ based on the parameters in $\A$. We also associate a uniform distribution over $\R(p;\A)$, assuming every value in $\R(p;\A)$ is equally likely to be realized if price $p$ is offered. Using this revenue distribution, we define a risk-adjusted revenue function, denoted $\eta(\cdotp;\A)$, as follows.
    \begin{definition}\label{def:risk-adjust}
        Given a nonempty ambiguity set $\A \subseteq \Theta$, the risk-adjusted revenue function with respect to $\A$ is the mapping $\eta(\cdot;\A): \pSet \to [0, \infty)$ that, for each price $p \in \pSet$, satisfies: (i) $\eta(p; \A) \geq \eta(p; \A')$ for every nonempty ambiguity set $\A' \subseteq \Theta$ such that $\A \subseteq \A'$, and (ii) $\eta(p; \A) = r(p; \theta^0)$ if $\A = \{\theta^0\}$.
    \end{definition}

    Condition (i) in Definition \ref{def:risk-adjust} ensures that the risk-adjusted revenue function becomes less conservative as the ambiguity set $\A$ shrinks toward $\{\theta^0\}$, and Condition (ii) requires it to coincide with the true expected revenue in the limiting case of $\A = \{\theta^0\}$. An example of this function is the $\alpha$-value at risk ($\alpha$-VaR). Fix a price $p \in \pSet$, an ambiguity set $\A\subseteq \Theta$, and a probability level $\alpha \in [0,1]$. Without loss of generality, assume the values in $\R(p;\A)$ are sorted from the smallest expected revenue $r_1$ to the largest $r_K$, where $K = |\R(p;\A)|$, so that $r_1 \le \cdots \le r_K$. The $\alpha$-VaR risk-adjusted revenue function is defined as $\etaVaR(p; \A) := r_{k_\alpha}$, where $k_\alpha := \max(1,\left\lceil \alpha K \right\rceil)$ denotes the index of the $k_\alpha$-th smallest value in $\R(p;\A)$. This definition includes a max-with-one operation to handle small values of $\alpha$ for which $\left\lceil \alpha K \right\rceil = 0$. We interpret the value of $r_{k_\alpha}$ as follows: if price $p$ is played, then the realized revenue is at least $r_{k_\alpha}$ with probability $1-\alpha$. For the zero probability level $\alpha = 0$, the $\alpha$-VaR risk-adjusted revenue function returns the the most conservative value of $\etaVaR(p; \A) = r_1 = \min \{ r : r \in \R(p;\A) \}$, which is a lower bound on the unknown realized revenue at price $p$ with probability one. For positive values of $\alpha$, the level of this conservatism decreases, allowing for a less pessimistic revenue assessment. More broadly, we can use other statistics of the uniform revenue distribution, such as the $\alpha$-conditional value at risk, to define alternative risk-adjusted revenue functions. \looseness=-1

\subsection{Partially Informative Prices}\label{sec:partially-inform-price} 

    Prices affect not only revenue but also the realized demand data, directly impacting how efficiently the true demand parameter $\theta^0$ is learned. Thus, prices should be selected not only to balance revenue and risk but also to enable the refinement of $\A$ toward $\{\theta^0\}$. To this end, we introduce partially informative prices.\looseness=-1
    \begin{definition}\label{def:risk-adjust-rev}
        A price $p\in\mathcal{P}$ is partially informative with respect to an ambiguity set $\A\subseteq \Theta$ if either (i) there is a pair of distinct parameters $(\theta^p,\bar \theta^p)\in\A^2$ satisfying $|\mu(p; \theta^p) - \mu(p; \bar \theta^p)| > 0$ or (ii) $\A = \{\theta^0\}$.
    \end{definition}
    To understand Definition \ref{def:risk-adjust-rev}, fix a price $p$ and an ambiguity set $\A$, and suppose Condition (i) is not met. In this case, all demand parameters $\theta \in \A$ yield the same mean demand $\mu(p; \theta)$, so playing price $p$ does not help distinguish between any pair of models in the ambiguity set. However, if Condition (i) is satisfied and sufficient data is collected at price $p$, then with high probability we should be able to statistically identify at least one parameter in $\A$ that differs from the true parameter $\theta^0$, and remove it from $\A$. In this case, price $p$ aids in learning $\theta^0$ and is thus called partially informative. Under Condition (ii), no further refinement is needed due to $\A = \{\theta^0\}$, so every price, including $p$, is partially informative. We define the set of all partially informative prices with respect to $\A \subseteq \Theta$, denoted $\PIP(\A)$, as $\PIP(\A) \equiv \mathcal{P}$ when $|\A| = 1$, and as follows when $|\A| \ge 2$:\looseness=-1
    \[
        \PIP(\A) := \big\{ p \in \mathcal{P} \ \big| \ \exists (\theta^p, \bar{\theta}^p) \in \A^2, \ \big|\mu(p; \theta^p) - \mu(p; \bar{\theta}^p)\big| > 0 \big\},
    \]
    Under Assumption \ref{asm:deman-model-set}, it is easy to verify that there is always at least one partially informative price in $\PIP(\A)$ for any ambiguity set $\A$, as shown in Proposition \ref*{prop:PI}.

    \setlength{\algomargin}{6pt}
    \begin{algorithm}[t]
        \DontPrintSemicolon
        \SetAlgoLined\vspace{6pt}
        \KwIn{%
        Parameter $\delta\in(0,1]$ and the risk-adjusted revenue function $\eta$.
        }
        \KwInit{%
        Ambiguity set $\hat\A_{1}\gets\Theta$ and dataset $\D(p)\gets\{\}$ for each $p\in\pSet(\Theta)$. 
        }
         \For{$t=1,2\dots,T$}{
                Play price $\hat{p}_t \gets \argmax\{\eta(p;\hat\A_t): p\in  \PIP(\hat\A_t)\}$ and update  dataset $\D(\hat p_t) \gets \D(\hat p_t)\cup\{\hat d_{i}:i \in\indexSet{N_t}\}$.\;
                Set $\hat\A_{t+1}\gets\hat\A_t$ if $|\D(\hat p_t)| < n_\delta(\hat p_t)$, else update $\hat\A_{t+1}\gets\bigl\{\theta\in\hat\A_t \big| \ell(\theta;\D(\hat p_t)) < c(\hat p_t)/2\bigr\}$.\;
                % =====> With two conditions
                % \If{$|\D(\hat p_t)|\ge n_\delta(\hat p_t)$ and $|\hat\A_t|\ge 2$}{
                % Update ambiguity set $\hat\A_{t+1} \gets \big\{\theta  \in \hat\A_t \ \big| \  \ell\big(\theta ;\D(\hat p_t)\big) <  c(\hat p_t) / 2  \big\}$.\;
                % }
                % \Else{
                %     Set $\hat\A_{t+1}\gets  \hat\A_t$.
                % }
         }\vspace{6pt}
        \KwOut{%
            The sequence of DAS  $(\hat\A_{1},\dots,\hat\A_{T})$ and prices $(\hat p_1, \dots, \hat p_T)$.
        }
        \caption{\normalfont{Adaptive Risk Learning}}
        \label{alg:ARL}
    \end{algorithm}

\subsection{Algorithm} \label{sec:alg}
    
    Algorithm \ref{alg:ARL} summarizes ARL, our joint pricing and demand learning framework. It takes as input the tunable parameter $\delta \in (0,1]$ and a risk-adjusted revenue function $\eta$ (e.g., $\alpha$-VaR). ARL maintains a data-driven ambiguity set (DAS), denoted by $\hat{\A}_t$ at time $t$, along with demand datasets $\{\D(p) : p \in \PIP(\Theta)\}$, where $\D(p)$ stores all data collected when price $p$ is played. At initialization, ARL sets $\hat{\A}_1 := \Theta$ to include all candidate demand parameters and initializes each $\D(p)$ to the empty set, as no data has been collected yet. At each time $t = 1,2,\dots,T$, ARL selects a partially informative price $\hat{p}_t \in \PIP(\hat{\A}_t)$ that maximizes the risk-adjusted revenue function $\eta(p; \hat{\A}_t)$. Upon offering $\hat{p}_t$, demand observations $\{\hat{d}_i : i \in \indexSet{N_t} \}$ are realized, where each $\hat{d}_i$ is an i.i.d. sample from $\distD(\hat{p}_t)$ (see \S\ref{sec:Problem Formulation}). These observations are added to the dataset $\D(\hat{p}_t)$. At time $t$, if the number of data points collected at price $\hat{p}_t$ (i.e., $|\D(\hat p_t)|$) is less than the threshold $n_\delta(\hat{p}_t)$, the DAS remains unchanged such that $\hat{\A}_{t+1} = \hat{\A}_t$ holds. This reflects the need for additional data to gain sufficient statistical confidence before refining the ambiguity set (see \S\ref{subsec:ARL-theory} for details). Otherwise, when the data threshold is met, ARL updates the DAS according to $\hat{\A}_{t+1} := \big\{ \theta \in \hat{\A}_t \mid \ell(\theta; \D(\hat{p}_t)) < c(\hat{p}_t)/2 \big\},$ retaining only those parameters $\theta \in \hat{\A}_t$ with loss below the threshold $c(\hat{p}_t)/2$. ARL outputs the full sequence of DAS $(\hat{\A}_1, \dots, \hat{\A}_T)$ and prices $(\hat{p}_1, \dots, \hat{p}_T)$. Importantly, ARL operates without requiring knowledge of the customer arrival pattern, which is given by the number of arrivals in each period.
    
    % =====> With two conditions
    % ARL updates the DAS if (i) the DAS contains more than one model (i.e., $|\hat{\A}_t| \geq 2$) and (ii) at least $n_\delta(\hat{p}_t)$ data points have been collected at price $\hat{p}_t$. If either condition fails, the DAS remains unchanged, i.e., $\hat{\A}_{t+1} = \hat{\A}_t$. If the first condition fails, ARL does not update the DAS because we can show that it essentially identified the true demand parameter $\theta^0$, and no further updates are needed. If the second condition fails, we can show that additional data is required to gain sufficient statistical confidence before refining the DAS. We provide these theoretical results in \S\ref{subsec:ARL-theory}. If both conditions hold, ARL updates the DAS to $\hat{\A}_{t+1} := \big\{ \theta \in \hat{\A}_t \mid \ell(\theta; \D(\hat{p}_t)) < c(\hat{p}_t)/2 \big\}$, retaining only those parameters $\theta$ from $\hat{\A}_t$ for which the loss value of $\ell(\theta; \D(\hat{p}_t))$ falls below the threshold $c(\hat{p}_t)/2$. ARL outputs the sequence of DAS $(\hat{\A}_1, \dots, \hat{\A}_T)$ and prices $(\hat{p}_1, \dots, \hat{p}_T)$. We now switch defining constant $c(\cdot)$, loss $\ell(\cdotp; \cdot)$, and integer $n_\delta(\cdot)$.

\begin{figure}[t]
    	\centering
    	\caption{Illustrating $c(p)$ based on four initial models (left) and updated DAS containing two models (right).}
        \includegraphics[width=1\linewidth]{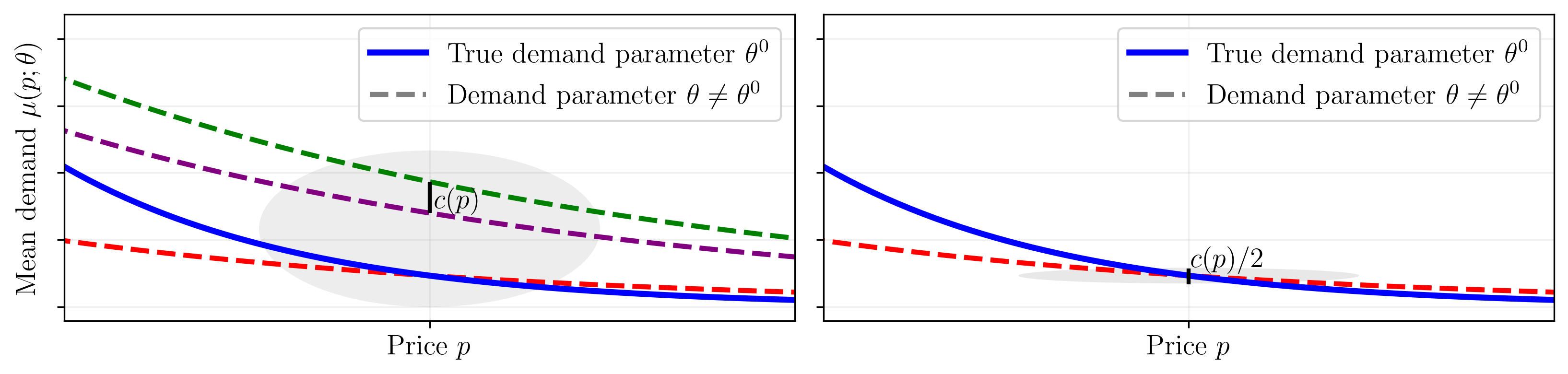}\vspace{-35pt}
        \label{fig:illustrating_c_p}
    \end{figure}

    The separation constant $c(p):= \min\big\{ |\mu(p; \theta) - \mu(p; \theta')| > 0 : (\theta, \theta') \in \Theta^2 \big\}$ is defined as the smallest non-zero difference between the mean demands of every two candidate demand models at a partially informative price $p \in \PIP(\Theta)$. This constant is well-defined under Assumption \ref{asm:deman-model-set}, which guarantees that for every pair $(\theta, \theta')$, there exists a price $p$ at which their expected demands differ. The left panel of Figure \ref{fig:illustrating_c_p} illustrates $c(p)$ when $\Theta$ includes four models. The solid blue curve shows the true mean demand $\mu(\cdot; \theta^0)$, while the dashed curves represent three other mean demands $\mu(\cdot; \theta)$ with parameters $\theta \ne \theta^0$. The vertical bracket labeled $c(p)$ highlights the smallest positive gap between any two distinct mean demand curves at price $p$. Smaller values of $c(p)$ suggest that there are two demand models within $\Theta$ with close means at price $p$, making discrimination between them more challenging at this price. This smaller value may also indicate a more challenging identification/learning of $\theta^0$.\looseness=-1

    The loss function $\ell(\theta;\D(p))$ for parameter $\theta \in \Theta$ and dataset $\D(p)$ at price $p \in \pSet$ is defined as follows:
    \[
        \ell(\theta;\D(p)) :=  \bigg|\mu\left(p; \theta\right) \ - \ \frac{1}{|\D(p)|} \sum_{d\in\D(p)} d \bigg|.
    \]
    It measures the absolute difference between the model-based mean demand $\mu(p; \theta)$ and the empirical mean $\frac{1}{|\D(p)|} \sum_{d \in \D(p)} d$. Smaller loss values suggest that the model with parameter $\theta$ is closer to the true demand model. This empirical mean is an unbiased estimator of $\mu(p; \theta^0)$, with its accuracy improving with the sample size $|\D(p)|$. To ensure sufficient statistical accuracy, Algorithm \ref{alg:ARL} requires $|\D(p)| \geq n_\delta(p)$, where for a confidence level $\delta \in (0,1]$ and price $p \in \PIP(\Theta)$, the threshold $n_\delta(p)$ is given by:
    \[
        n_\delta(p) := 4 \cdot \max\left\{2\left(\frac{v}{c(p)}\right)^2, \frac{b}{c(p)}\right\} \cdot \ln\left(\frac{2}{\delta}\right),
    \]
    where the sub-exponential constants $v$ and $b$ are defined in Assumption \ref{assump:LightTailedDemand}.
 
    The definitions of the separation constant $c(\cdot)$, loss function $\ell(\cdot;\D(\cdot))$, and threshold $n_\delta(\cdot)$ are interconnected through a concentration inequality that plays a central role in our analysis. This inequality depends on the set $\I(p) := \big\{\theta \in \Theta : \mu(p; \theta) = \mu(p; \theta^0)\big\}$ that contains all demand parameters in $\Theta$ (including $\theta^0$) whose mean demand equals the true mean at price $p \in \pSet$. For instance, in the left panel of Figure \ref{fig:illustrating_c_p}, the red dashed curve intersects the blue solid curve (true mean) at price $p$, indicating that both models lie in $\I(p)$. This concentration inequality at iteration $t$ of Algorithm \ref{alg:ARL}, where inequality $|\D(\hat{p}_t)| \ge n_\delta(\hat{p}_t)$ holds, is given by:
    \[
        \prob\left( \max_{\theta \in \I(\hat{p}_t)} \ell(\theta; \D(\hat{p}_t)) \le \frac{c(\hat{p}_t)}{2}, \ \max_{\theta \in \Theta \backslash \I(\hat{p}_t)} \ell(\theta; \D(\hat{p}_t)) > \frac{c(\hat{p}_t)}{2} \right) \ge 1 - \delta,
    \]
    where the probability operator $\prob(\cdot)$ is over the randomness in price $\hat{p}_t$, given DAS $\hat{\A}_t$ and dataset $\D(\hat{p}_t)$. This inequality implies that, with probability at least $1 - \delta$, all intersecting models $\theta \in \I(\hat{p}_t)$ have loss at most $c(\hat{p}_t)/2$, while all non-intersecting models $\theta \in \Theta \backslash \I(\hat{p}_t)$ have loss strictly greater than $c(\hat{p}_t)/2$. Hence, under the DAS update rule, the updated set $\hat{\A}_{t+1}$ includes all intersecting models with probability at least $1 - \delta$. As the parameter $\delta \in (0,1]$ decreases to zero, this probability level increases to one, which in turn raises the threshold $n_\delta(\hat{p}_t)$, requiring more data to be collected at price $\hat{p}_t$ before updating the DAS. The right panel of Figure \ref{fig:illustrating_c_p} illustrates the updated DAS for the example in the left panel, where the initial DAS (left) includes four models, while the updated DAS (right) includes only the two models intersecting at price $p$.\looseness=-1
 
\subsection{Theoretical Results}\label{subsec:ARL-theory}
    
    In this section, we establish the convergence of the DAS and develop a regret bound for the ARL policy. The discussion in \S\ref{sec:alg} and Figure \ref{fig:illustrating_c_p} suggest that the DAS at each time should include those demand parameters whose mean demands intersect at the prices selected by ARL. We formalize this intuition in Theorem \ref{thm:DAS-evolution}.\looseness=-1
    \begin{theorem}\label{thm:DAS-evolution}
        Fix $\delta \in (0,1]$. For every sequence $(\hat\A_{1}, \dots, \hat\A_{T})$ and $(\hat{p}_{1}, \dots, \hat{p}_{T})$ obtain from Algorithm \ref{alg:ARL}, at each time $t \in \indexSet{T}$, with probability of $1 - \delta$, the inclusion property $\theta^0 \in \hat\A_{t}$ and the following identity hold:\looseness=-1
        \begin{equation}\label{eq:update_U}
            \hat{\A}_{t+1} = 
            \begin{dcases}
                \hat{\A}_{t} & \text{if } |\D(\hat{p}_t)| < n_\delta(\hat{p}_t), \\
                \hat{\A}_{t} \cap \I(\hat{p}_t) & \text{otherwise}.
            \end{dcases}
        \end{equation}
    \end{theorem}
    Theorem \ref{thm:DAS-evolution} shows that the sequence of DAS $(\hat{\A}_{1}, \dots, \hat{\A}_{T})$ generated by ARL shrinks over time according to the intersecting models while ensuring, with high probability, that the true demand model is never eliminated. Initially, at $t = 1$, the DAS is $\hat{\A}_1 = \Theta$, reflecting high demand model ambiguity. At $t = 2$, if the minimum-data condition $|\D(\hat{p}_1)| \ge n_\delta(\hat{p}_1)$ is satisfied, the DAS contracts to $\hat{\A}_2 = \I(\hat{p}_1) \subsetneq \Theta$, indicating reduced ambiguity. Otherwise, the DAS remains unchanged, i.e., $\hat{\A}_2 = \hat{\A}_1$, signaling that not enough data has been collected to reduce demand model ambiguity with sufficient statistical confidence. Thus, ARL dynamically shrinks DAS by adapting the level of demand model ambiguity to the unknown number of arriving customers.\looseness=-1

    This nested refinement of the DAS in Theorem \ref{thm:DAS-evolution} also suggests that $\theta^0$ should ultimately be the only parameter remaining in the DAS. Specifically, $\theta^0$ is the only demand parameter that belongs to the intersection of parameters whose mean demands coincide with that of the true model at all partially informative prices, i.e., $\bigcap_{p \in \PIP(\Theta)} \I(p) = \{\theta^0\}$. To see why, suppose there exists an alternative parameter $\theta \ne \theta^0$ in this intersection. Then, by Assumption \ref{asm:deman-model-set}, there must exist a partially informative price $p$ at which $\mu(p; \theta) \ne \mu(p; \theta^0)$. However, this contradicts the assumption that $\theta \in \I(p)$ for all $p \in \PIP(\Theta)$, suggesting such an alternative parameter cannot exist.  Therefore, in the worst case, if ARL plays all prices in $\PIP(\Theta)$ and encounters $n_\delta(p)$ samples at each, the DAS should intuitively contracts to $\{\theta^0\}$ with high probability. 
    
    We refer to the first time that the DAS contains only $\theta^0$ as the demand identification time. To elaborate, consider an environment in which the mean demands $\{\mu(p;\theta) : \theta \in \Theta\}$ are non-intersecting at the price $\hat{p}_1 = \argmax\{\eta(p;\Theta) : p \in \PIP(\Theta)\}$. ARL continuously plays this price from $t = 1$ until some time $t \ge 1$ when the number of data points $|\D(\hat{p}_1)| = \sum_{s=1}^{t} N_s$ exceeds the threshold $n_\delta(\hat{p}_1)$. At that point, the DAS shrinks from $\Theta$ to $\{\theta^0\}$ according to \eqref{eq:update_U}, noting that $\I(\hat{p}_1)$ contains only $\theta^0$. Thus, in this environment, the demand identification time, denoted $\tau(\delta)$, is the first time $t$ at which condition $\sum_{s=1}^{t} N_s \ge n_\delta(\hat{p}_1)$ is met for a given $\delta\in(0,1]$. If the condition is not satisfied even at the last time $t = T$, we conclude that there is not enough data for ARL to identify $\theta^0$ with high probability. 

    To generalize the demand identification time to environments where demand models intersect, we define a deterministic analogue of the DAS as follows. Initialize $\bar{\A}_1 := \Theta$. For each integer $k \ge 1$, define $\bar{p}_k := \argmax\{\eta(p; \bar{\A}_k) : p \in \PIP(\bar{\A}_k)\}$ and update $\bar{\A}_{k+1} := \bar{\A}_k \cap \I(\bar{p}_k)$. Let $K$ be the smallest index for which $\bar{\A}_{K+1} = \{\theta^0\}$ holds. To connect the ambiguity set $\bar{\A}_k$ to the DAS $\hat\A_t$, initialize $\bar{t}_0(\delta) := 0$ and define $\bar{t}_{k}(\delta)$ for each $k = 1, 2, \dots, K$ as follows:
    \[
        \bar{t}_k(\delta) := \min\bigg\{ \min\bigg\{t \in \{\bar{t}_{k-1}(\delta)+1, \dots, T\} : \sum_{s=\bar{t}_{k-1}(\delta)+1}^{t} N_s \ge n_\delta(\bar{p}_k) \bigg\},\, T+1 \bigg\}.
    \]
    The value of $\bar{t}_1(\delta)$ can be interpreted as the earliest period in ARL at which enough data have been gathered to shrink $\hat{\A}_1 = \Theta$ to $\hat{\A}_2 = \I(\hat{p}_1)$ with probability $1-\delta$. Similarly, $\bar{t}_2(\delta)$ corresponds to the first period in which the ambiguity set reduces to $\hat{\A}_3 = \I(\hat{p}_1) \cap \I(\hat{p}_2)$, and so forth. We can now define the demand identification time, which is the earliest time at which the ambiguity set $\Theta$ shrinks to $\bar{\A}_{K+1} = \{\theta^0\}$, as follows:\looseness=-1\vspace{-12pt}
    \[
        \tau(\delta) := \bar{t}_{K}(\delta).\vspace{-12pt}
    \]

    Proposition \ref{prop:stoch-to-det} links the DAS $\hat\A_t$ to its deterministic analogue $\bar{\A}_k$ and shows its convergence to $\{\theta^0\}$ with high probability after the demand identification time $t \ge \tau(\delta)+1$.
    \begin{proposition}\label{prop:stoch-to-det}
        Fix $\delta \in (0,1]$. For every sequence of DAS $(\hat\A_{1}, \dots, \hat\A_{T})$ obtained from Algorithm \ref{alg:ARL} and for each $k \in \indexSet{K}$, the following holds:
        \[
            \prob\big(\hat\A_t=\bar\A_k\big)\ge 1-(k-1)\delta, \quad  \forall t\in\big\{\bar t_{k-1}(\delta)+1,\dots,\bar t_k(\delta)\big\}. 
        \]
        % the identity $\hat{\A}_{t} = \bar{\A}_{k}$ holds for $t \in \{\bar{t}_{k-1}(\delta)+1, \bar{t}_{k-1}(\delta)+2, \dots, \bar{t}_{k}(\delta)\}$ with probability at least $1 - \delta$ . 
        Moreover, for every time $t \in \{\tau(\delta)+1, \dots, T\}$, we have $\hat{\A}_{t} = \{\theta^0\}$ with probability $1-K\delta$.
    \end{proposition} 
    To understand Proposition \ref{prop:stoch-to-det}, recall that ARL in Algorithm \ref{alg:ARL} updates the DAS only at specific iterations when the else-condition is triggered. Each such iteration corresponds to an index $k$ in the proposition, and the evolution of the DAS is tracked by the deterministic set $\bar{\A}_k$. For example, when $k = 1$, the DAS remains unchanged as $\hat\A_t = \Theta$ for all $t \in \{1, \dots, \bar{t}_1(\delta)\}$. Time $t = \bar{t}_1(\delta)$ marks the first refinement of the DAS to a smaller set, i.e., $\hat\A_{t+1} \subsetneq \hat\A_t$. Thereafter, the DAS remains fixed for all $t \in \{\bar{t}_1(\delta)+1, \dots, \bar{t}_2(\delta)\}$, and this pattern continues for subsequent updates. Proposition \ref{prop:stoch-to-det} also shows that once enough data are collected, the DAS converges to the singleton $\{\theta^0\}$ with high probability at the demand identification time. Together with Theorem \ref{thm:DAS-evolution}, this result implies that ARL tracks demand model ambiguity through the DAS and adaptively refines it over time as more data become available, eventually isolating the true demand parameter.\looseness=-1

    These theoretical results on the convergence of the DAS reveal a key insight: ARL smoothly transitions from risk aversion, when ambiguity is high and the DAS includes all demand model parameters, to revenue maximization, as ambiguity diminishes and the DAS narrows to only the true parameter. This transition is directly tied to the regret of the ARL policy. Early on, when the knowledge of $\theta^0$ is limited, ARL may incur suboptimal revenues by prioritizing risk mitigation. As more data are collected, ambiguity decreases, and the policy shifts toward revenue maximization by refining the DAS to isolate $\theta^0$. Intuitively, the timing of this shift, captured by the demand identification time, should govern the ARL policy's regret. Denote by $\piARL$  the ARL policy, whose time-$t$ decision rule $\piARL_t$ selects the price $\hat{p}_t$ in Algorithm~\ref{alg:ARL} based on the current DAS $\hat{\A}_t$ and the observed history of past prices and demand realizations. The regret for the ARL policy is defined as:\looseness=-1
    \[
        R_T(\piARL) := \CIR(\piCI) - \CIR(\piARL),
    \]
    which quantifies the difference in expected cumulative revenues between the clairvoyant CI policy that knows $\theta^0$ and the ARL policy that learns it from data. Theorem \ref{thm:regret} bounds this regret in terms of the demand identification time and the constant $\bar{r} := \max\big\{ r(p; \theta^0): p\in\pSet \big\}$, maximum expected revenue under $\theta^0$ across all prices.\looseness=-1
    \begin{theorem}\label{thm:regret}
        Fix $\delta \in (0,1]$. With probability $1 - K\delta$, the following regret bound holds: 
        \[
            R_T(\piARL) \ \le \ 2\bar{r}\sum_{t=1}^{\min(\tau(\delta),T)} N_t.
        \]
    \end{theorem}

    Theorem \ref{thm:regret} bounds the ARL policy's regret by $2\bar{r}$ times the total number of customer arrivals up to $t=\min\{\tau(\delta), T\}$. If the true demand model is identified within the $T$‐period horizon, that is, when $\tau(\delta)\le T$, regret scales with the arrivals that occur before the demand identification time. Otherwise, when identification is not achieved within $T$ periods (so $\tau(\delta)=T+1$), the regret bound equals $2\bar{r}M$, where $M$ is the total store traffic. For instance, under the worst‐case customer arrival pattern $(N_1,N_2,\dots,N_T)=(M,0,\dots,0)$, in which all customers arrive in period 1, we have $\tau(\delta)=T+1$, yielding a regret bound linearly increasing in $M$. In more favorable patterns, where a significant share of customers arrives after identification, the ARL policy can achieve stronger regret, sublinear in $M$. Thus, Theorem \ref{thm:regret} explicitly captures the impact of customer arrival patterns on the performance of the ARL policy, a direct consequence of permitting only infrequent price changes in our constrained dynamic pricing setting.\looseness=-1
    
    An important special case is the commonly studied unconstrained dynamic pricing setting, where the firm observes one customer per period  such that $N_t = 1$ for every $t \in \indexSet{T}$ and $M = T$. Because the firm can update the price after each arrival, a suboptimal price can be revised quickly and is not offered to many customers, expecting low regret for reasonable policies. We show that the ARL policy attains this performance because the policy offers each partially informative price $\hat{p}_t$ to exactly $n_\delta(\hat{p}_t)$ customers, after which it updates this price since it has obtained sufficient statistical confidence to shrink the DAS. Corollary \ref{cor:unconstrained} shows that the ARL policy in the unconstrained dynamic pricing environment incurs a constant regret.
    \begin{corollary}\label{cor:unconstrained}
        Fix $\delta \in (0,1]$. In the unconstrained dynamic pricing setting with $N_t = 1$ for each $t \in \indexSet{T}$, the regret bound below holds with probability $1 - K\delta$:
        \[
             R_T(\piARL) \ \le \ 2\bar{r} \cdot \min\left\{\sum_{k=1}^K n_\delta(\bar{p}_{k}), \  T\right\}.
        \]
    \end{corollary}
    The regret bound in Corollary \ref{cor:unconstrained} replaces the arrival-dependent term $\sum_{t=1}^{\min(\tau(\delta),T)} N_t$ in Theorem \ref{thm:regret} with the arrival-agnostic constant $2\bar{r}\cdot\min\{\bar{n}_\delta, T\}$, where $\bar{n}_\delta := \sum_{k=1}^K n_\delta(\bar{p}_{k})$. Thus, in the unconstrained setting, the ARL policy achieves a constant regret bound, independent of both $M$ and $T$, such that $\lim_{T\rightarrow\infty} R_T(\pi_{\text{ARL}}) = 2\bar{r}\cdot\bar{n}_\delta$ holds with probability $1 - K\delta$. This result builds on Proposition 1 of \cite{CSW2017}, which establishes a constant regret bound for a pricing policy, but differs in two key ways. First, their policy changes prices infrequently but allows an unbounded number of updates, whereas ARL changes prices only a finite number of times. Second, their guarantee relies on the existence of a discriminative price (Assumption 1 in \citealt{CSW2017}), which ensures separability of all demand models at this price point. Our bound requires no such separability assumption. Thus, our work extends these known results to settings with only a few price changes and without the assumption of separability. Interestingly, we can interpret $\bar{n}_\delta$ as a condition number: as the candidate demand models become harder to distinguish, the separation constants $c(\bar{p}_k)$ decrease, causing each $n_\delta(\bar{p}_k)$, and hence $\bar{n}_\delta$, to increase, reflecting a more challenging demand identification.\looseness=-1
        
\section{The Value of Adaptation}\label{sec:value-of-adaptation}

    In this section, we explore the benefits of ARL’s adaptation mechanism for risk mitigation and demand learning. We quantify the value of adapting the level of risk aversion to the customer arrival pattern via the DAS, by considering a non-adaptive risk-mitigating (NRM) policy that replaces the adaptive data-driven DAS $\hat\A_t$ in ARL with the fixed data-agnostic set $\Theta$. NRM preserves ARL's risk-mitigation objective but removes adaptivity, allowing us to isolate the benefit of ARL's data-driven risk adjustment. To uncover the value of DAS in demand learning, we define a follow-the-leader (FTL) policy that replaces ARL’s set estimate $\hat\A_t$ of $\theta^0$ with the point estimate $\thetaFTL{t}\in\Theta$. Because FTL focuses on learning rather than risk mitigation, comparing it with ARL enables us to isolate the value of demand learning via set estimation in ARL. Consider the following objective functions:\looseness=-1
    \begin{align*}
        \ARLO(p; \hat\A_{t}) &:=\ \eta(p; \hat\A_{t}) \\
        \NRM(p; \Theta) &:=\ \eta(p; \Theta) \\
        \FTLO(p; \thetaFTL{t}) &:=\ r(p; \thetaFTL{t}) \\
        \CIO(p; \theta^0) &:=\ r(p; \theta^0)
    \end{align*}
    These objectives are optimized by the ARL, NRM, FTL, and CI policies, respectively. The argument after the semicolon in each function specifies the information used to evaluate it. In \S\ref{sec:NRM}, we compare the ARL and NRM policies to quantify ARL’s benefit from adapting the ambiguity set $\Theta$ to data. In \S\ref{sec:FTL}, we compare the ARL and FTL policies to assess ARL’s benefit of learning $\theta^0$ via adaptive, data-driven ambiguity sets.\looseness=-1

\subsection{Risk Mitigation: Adaptive vs Non-Adaptive}\label{sec:NRM}

    The risk-focused NRM policy mitigates potential revenue loss due to demand model ambiguity by selecting a price that maximizes $\NRM(p; \Theta)$, the risk-adjusted revenue function with respect to the set of all candidate parameters $\Theta$. Denoting the NRM policy by $\piNRM$, it applies the fixed price $\pNRM \in \pSet$ at every stage $t$, where $\pNRM$ solves the optimization problem $\max\{\eta(p; \Theta):p \in \pSet\}$. Since the NRM policy does not focus on learning, its price is optimized over the entire set $\pSet$ rather than being restricted to a partially informative subset, which is needed for learning, as explained in \S\ref{sec:partially-inform-price}. To interpret the NRM price $\pNRM$, consider the $\alpha$-VaR risk-adjusted revenue function $\etaVaR(\cdotp; \Theta)$ from \S\ref{sec:risk-adjust-rev}. If $\alpha = 0$, this function returns the worst-case revenue in the set $\R(p;\Theta)$ defined in \S\ref{sec:risk-adjust-rev}, so $\pNRM$ solves the maximin problem $\max_{p \in \pSet} \min_{\theta \in \Theta} r(p; \theta)$. In this case, $\pNRM$ can be viewed as the ``robust'' price maximizing a worst revenue. Proposition \ref{prop:rev-ARL-NRM} provides a comparison between the NRM and ARL revenue functions relative to the clairvoyant CI revenue function $\CIO(\cdotp; \theta^0)$.\looseness=-1
    
    \begin{proposition}[Adaptation Tightens Conservative Revenue Assessments]\label{prop:rev-ARL-NRM}
        Fix $\delta \in (0,1]$. For each sequence of DAS $(\hat\A_{1}, \dots, \hat\A_{T})$ obtained from Algorithm \ref{alg:ARL}, the following holds with probability $1 - \delta$:\looseness=-1
        \begin{equation*}
            \big|\CIO(p; \theta^0) - \ARLO(p; \hat\A_t)\big| \ \le \ \big|\CIO(p; \theta^0) - \NRM(p; \Theta)\big|, \quad \forall p \in \pSet, \ \forall t\in\indexSet{T}.
        \end{equation*}
    \end{proposition}
    
    Proposition \ref{prop:rev-ARL-NRM} shows that the ARL revenue function is closer to the unknown CI revenue function with high probability than the NRM revenue function. This follows from the probabilistic inclusion $\{\theta^0\} \subseteq \hat\A_t \subseteq \Theta$, where $\theta^0$, $\hat\A_t$, and $\Theta$ are used in the CI, ARL, and NRM revenue functions, respectively. In early periods, when model ambiguity is high and ARL has not yet updated the DAS, we expect $\hat\A_t = \Theta$, leading ARL and NRM to yield the same conservative revenue assessments. As data accumulates and the DAS shrinks, the strict inclusion $\hat\A_t \subsetneq \Theta$ emerges, making the ARL revenue less conservative than that of NRM. After the demand identification time, the identity $\hat{\A}_{t} = \{\theta^0\}$ holds with high probability by Proposition \ref{prop:stoch-to-det}, implying ARL and CI revenue functions match, i.e., $\ARLO(p; \hat\A_{t}) = \CIO(p; \theta^0)$. This equality does not hold in general for the NRM objective at any time, since $\Theta$ remains agnostic to data and contains models beyond $\theta^0$. Thus, the adaptation mechanism in ARL reduces the over-conservative revenue assessments of NRM over time, eventually aligning them with the CI revenue function upon the convergence of the DAS.

    The non-adaptive nature of the NRM policy suggests that its regret should grow linearly with $M$, independent of the customer arrival pattern. Proposition \ref*{prop:regret-NRM} formalizes this observation by bounding the NRM policy's regret $R_T(\piNRM)$ above by the constant $\bar{r}\cdot M$, assuming the CI and NRM prices are distinct (i.e., $\pCI \ne \pNRM$), which holds in general. This bound’s insensitivity to the customer arrival pattern stems from the fact that the NRM policy does not update the ambiguity set $\Theta$ based on data. In contrast, the regret bound for the ARL policy in Theorem \ref{thm:regret} explicitly depends on the arrival pattern. Under the worst-case scenario, where all customers arrive in the first period (i.e., $N_1 = M$), ARL incurs linear regret, similar to NRM. However, in more favorable settings, ARL can achieve sublinear regret, which is not possible under the NRM policy. Collectively, propositions \ref{prop:rev-ARL-NRM} and \ref*{prop:regret-NRM} demonstrate that the adaptation mechanism in ARL strengthens both revenue assessments and regret guarantees relative to the NRM policy.

\subsection{Demand Learning: Point Estimation vs Set Estimation}\label{sec:FTL}

    There are different ways to construct a follow-the-leader learning algorithm tailored to our constrained dynamic pricing setting. We present a version, detailed in Algorithm \ref{alg:FTL}, which mimics the structure of ARL but replaces the set estimation of $\theta^0$ via the DAS $\hat{\A}_t \subseteq \Theta$ with a point estimate $\thetaFTL{t} \in \Theta$. Because Algorithm \ref{alg:FTL} follows similar steps to Algorithm \ref{alg:ARL}, we do not repeat them here and only discuss the differences. First, FTL takes as input an initial guess $\thetaFTL{1}$ for the true parameter $\theta^0$, which is not required by ARL. Next, FTL neither optimizes a risk-adjusted revenue function nor maintains an ambiguity set. Specifically, at each time $t \in \indexSet{T}$, FTL prescribes a price $\pFTL_t \in \pSet$ that maximizes the expected revenue under the current estimate $\thetaFTL{t}$ (i.e., $\pFTL_t \in \argmax\{r(p; \thetaFTL{t}) : p \in \pSet\}$). The price $\pFTL_t$ is chosen from the full set of price candidates $\pSet$, rather than a partially informative subset, as such prices do not directly help FTL learn the true demand parameter faster. Finally, FTL updates its best estimate $\thetaFTL{t+1}$ by solving the supervised learning problem $\min_{\theta \in \Theta} \ell(\theta; \D(\pFTL_t))$ only when the number of data points collected at price $\pFTL_t$ exceeds the threshold $n_\delta(\pFTL_t)$, mimicking the DAS updates in ARL. FTL returns the sequence of point estimates $(\thetaFTL{1}, \dots, \thetaFTL{T})$ and prices $(\pFTL_1, \dots, \pFTL_T)$.

    \setlength{\algomargin}{6pt}
    \begin{algorithm}[t]
        \DontPrintSemicolon
        \SetAlgoLined\vspace{6pt}
        \KwIn{%
        Parameter $\delta\in(0,1]$ and initial estimate $\thetaFTL{1}$. 
        }
        \KwInit{%
        Dataset $\D(p) \gets \{\}$ for each $p \in \pSet$.
        }
         \For{$t=1,2\dots,T$}{
                Play price $\pFTL_t  \gets \argmax\{r(p;\thetaFTL{t}): p\in  \pSet\}$ and update dataset $\D(\pFTL_t ) \gets \D(\pFTL_t )\cup\{\dFTL_{i}:i \in\indexSet{N_t}\}$.\;
                Set $\thetaFTL{t+1}  \gets  \thetaFTL{t}$ if $|\D(\pFTL_t)| < n_\delta(\pFTL_t)$, else update $\thetaFTL{t+1} \gets \argmin\big\{\ell\big(\theta; \D(\pFTL_t)\big):\theta \in \Theta\big\}$.\;
                % \If{$|\D(\pFTL_t)|\ge n_\delta(\pFTL_t)$}{
                % Update parameter: $\thetaFTL{t+1} \gets \argmin_{\theta \in \Theta} \ell\big(\theta; \D(\pFTL_t)\big)$. \;
                % }
                % \Else{
                %     Set $\thetaFTL{t+1}  \gets  \thetaFTL{t}$.
                % }
         }\vspace{6pt}
        \KwOut{%
            The sequence of estimates $(\thetaFTL{1}, \dots, \thetaFTL{T})$ and prices $(\pFTL_1, \dots, \pFTL_T)$
        }
        \caption{\normalfont{Follow-the-Leader}}
        \label{alg:FTL}
    \end{algorithm}

    Proposition \ref{prop:rev-risk-FTL} provides an easy-to-verify comparison between the worst-case revenue assessments under the FTL and ARL policies and relates them to the unknown CI revenue function.
    \begin{proposition}[Set Estimation Tightens Worst-Case Revenue Assessments]\label{prop:rev-risk-FTL}% 
        For the CI, ARL, and FTL revenue functions, the following inequalities hold:\looseness=-1
        \begin{equation*}
            \CIO(p; \theta^0) \ \ge \ \min_{\{\theta^0\}\subseteq\A\subseteq\Theta}\ARLO(p;\A) \ \ge \ \min_{\theta\in\Theta}\FTLO(p;  \theta),\qquad \forall p\in\pSet, \ \forall t\in\indexSet{T}. 
        \end{equation*}
    \end{proposition}
    Proposition \ref{prop:rev-risk-FTL} shows that for every time $t \in \indexSet{T}$, the worst-case ARL revenue $\min\{\ARLO(p;\A) : \{\theta^0\} \subseteq \A \subseteq \Theta\}$ is no smaller than the worst-case FTL revenue $\min_{\theta \in \Theta} \FTLO(p; \theta)$. It also establishes that the worst-case ARL revenue is closer to the CI revenue $\CIO$ than the worst-case FTL revenue. These results highlight that if we repeatedly run ARL and FTL (in algorithms \ref{alg:ARL} and \ref{alg:FTL}) and compare their time-$t$ revenue functions, the worst-case ARL revenue across runs is not worse than that of FTL. Therefore, the revenue assessment in ARL based on set estimation reduces the risk of obtaining loose revenue assessments from FTL.\looseness=-1

    Proposition \ref{prop:FTL-param} establishes a key property of FTL's point estimator $\thetaFTL{t}$.
    \begin{proposition}\label{prop:FTL-param}
        Fix $\delta \in (0,1]$. If at time $t \in \indexSet{T}$ the else-condition in FTL occurs, then the inclusion $\thetaFTL{t+1} \in \I(\pFTL_t)$ holds with probability $1 - \delta$.
    \end{proposition}
    Proposition \ref{prop:FTL-param} shows that, whenever the else-condition in FTL is triggered, the updated FTL estimate $\thetaFTL{t+1}$ lies in the set of intersecting models at the FTL price $\pFTL_t$, namely $\I(\pFTL_t)$, with probability $1-\delta$. If this set contains only the true demand parameter, so that $\I(\pFTL_t) = \{\theta^0\}$, then FTL has essentially learned $\theta^0$ at time $t$ with high probability, since $\thetaFTL{t+1} \in \I(\pFTL_t) = \{\theta^0\}$. However, this assumption is restrictive because multiple demand models can intersect at the FTL price $\pFTL_t$. Thus, Proposition \ref{prop:FTL-param} shows that FTL can separate non-intersecting models with parameters in $\Theta\setminus\I(\pFTL_t)$ from intersecting ones with parameters in $\I(\pFTL_t)$, but it does not necessarily distinguish $\theta^0$ from other parameters within $\I(\pFTL_t)$ itself. This lack of learning directly impacts the FTL policy's regret. When $\I(\pFTL_t)$ contains multiple candidates, so that there exists a parameter $\theta' \in \I(\pFTL_t)$ with $\theta' \ne \theta^0$, Proposition \ref{prop:FTL-param} suggests that FTL may incorrectly select $\theta'$ rather than $\theta^0$ as its next estimate, making $\thetaFTL{t+1} = \theta'$ plausible. Therefore, if the subsequent FTL price $\pFTL_{t+1}$, optimized under the misspecified parameter $\theta'$, yields revenue substantially below the CI revenue (i.e., $r(\pFTL_{t+1};\theta^0)\ll r(\pCI;\theta^0)$), the FTL policy’s regret can become large, especially when many customers arrive at time $t+1$ and face this suboptimal price $\pFTL_{t+1}$. Overall, Proposition \ref{prop:FTL-param} suggests that the FTL policy can incur high regret when candidate demand models intersect, but can achieve strong regret performance under the restrictive assumption that models do not intersect at any price level, as formalized in Assumption \ref{asm:separability}.\looseness=-1
    \begin{assumption}\label{asm:separability}
        There exists a positive constant $\hat{c} > 0$ such that for every pair of distinct demand parameters $\theta, \theta' \in \Theta$, the inequality $|\mu(p; \theta) - \mu(p; \theta')| \ge \hat{c}$ holds at all prices $p \in \pSet$.
    \end{assumption}

    Under Assumption \ref{asm:separability}, we have $\I(p)=\{\theta^0\}$ for every $p \in \pSet$. Combined with Proposition \ref{prop:FTL-param}, this implies that $\thetaFTL{t+1} = \theta^0$ with probability $1-\delta$ whenever the else-condition in FTL occurs. Thus, FTL should learn $\theta^0$ once it collects enough data to update its initial estimate $\thetaFTL{0}$ to $\thetaFTL{1}$. To formalize this idea, recall that $\bar{t}_1(\delta)$ denotes the earliest period in ARL at which sufficient data have been gathered to shrink $\hat{\A}_1=\Theta$ to $\hat{\A}_2=\I(\hat{p}_1)$ with probability $1-\delta$. Under Assumption \ref{asm:separability}, we have $\hat{\A}_2=\{\theta^0\}$, implying that the smallest index $K$ satisfying $\bar{\A}_{K+1}=\{\theta^0\}$ is $K=1$. Accordingly, the demand identification time under this assumption becomes $\tau(\delta)=\bar{t}_{K}(\delta)=\bar{t}_1(\delta)$. The demand identification time tailored to Assumption~\ref{asm:separability} can be rewritten as follows:\looseness=-1
    \[
         \tau(\delta):= \bar{t}_1(\delta) = \min\bigg\{ \min\bigg\{t \in \{1, \dots, T\} : \sum_{s=1}^{t} N_s \ge 4 \cdot \max\left\{2\left(\frac{v}{\hat{c}}\right)^2, \frac{b}{\hat{c}}\right\} \cdot \ln\left(\frac{2}{\delta}\right) \bigg\}, \ \   T+1 \bigg\},\vspace{-2mm}
    \]
    noting that we replaced the price-dependent separation constant $c(p)$ with the uniform separation constant $\hat{c}$. In Theorem \ref{thm:regret-FTL}, we relate the demand learning and regret properties of ARL and FTL under Assumption \ref{asm:separability}.

    \begin{theorem}[Point Estimation Requires Separability for Learning and Regret Guarantees]\label{thm:regret-FTL} 
        Fix $\delta \in (0,1]$ and require Assumption \ref{asm:separability}. For every $t \ge \tau(\delta)$, the identity $\hat{\A}_t = \{\thetaFTL{t}\} = \{\theta^0\}$ holds with probability $1 - \delta$. Moreover, with the same probability, the FTL policy's regret satisfies:
        \[
            R_T(\piFTL) \ \le \ 2\bar{r}\sum_{t=1}^{\min(\tau(\delta),T)} N_t.
        \]
    \end{theorem}
    
    Theorem \ref{thm:regret-FTL} shows that both FTL and ARL learn the true demand parameter $\theta^0$ with high probability under Assumption \ref{asm:separability} as soon as the demand identification time is reached. Under this assumption, FTL also attains an arrival-dependent regret bound that scales with the number of customers arriving before identification. This bound coincides with the ARL regret bound in Theorem \ref{thm:regret}, since Assumption \ref{asm:separability} implies $K=1$ and both theorems yield the same bound, $2\bar{r}\sum_{t=1}^{\min(\tau(\delta),T)} N_t$, with probability $1-\delta$. Thus, the insights in \S\ref{subsec:ARL-theory} on how ARL regret depends on the arrival pattern directly carry over to FTL under Assumption \ref{asm:separability}. Theorem \ref{thm:regret-FTL} also highlights that when demand models are fully separated, a restrictive condition, set estimation in ARL is not needed for learning, as FTL also recovers $\theta^0$. Even in this case, however, ARL retains its key advantage of mitigating the risk of revenue loss, a feature absent in FTL. More importantly, when Assumption \ref{asm:separability} fails, which is the typical setting, FTL may not identify $\theta^0$ and can incur large regret, whereas ARL provides both learning and regret guarantees. Overall, generalizing the point-estimation mechanism in FTL to incorporate richer information through set estimation yields two central benefits for ARL: (i) stronger risk mitigation, as shown in Proposition \ref{prop:rev-risk-FTL}, and (ii) guaranteed convergence to the true demand model without requiring separability.\looseness=-1
        \section{Computational Study}\label{sec:numbers}

    In this section, we evaluate the numerical performance of the ARL policy. In \S\ref{sec:Instances}, we outline our test instances. In \S\ref{sec:implementation}, we describe the implementation details of the ARL, FTL, and NRM policies. In \S\ref{sec:results}, we report our numerical results and highlight the benefits of the adaptation mechanism of ARL.

\subsection{Problem Instances}\label{sec:Instances}

    We consider 270 test instances of the constrained dynamic pricing problem introduced in \S\ref{sec:Problem Formulation}. We vary the mean demand $\mu(p;\theta)$, the additive noise $\boldsymbol{\zeta}$, the candidate parameter set $\Theta$, and the arrival pattern $(N_1,\ldots,N_T)$. Mean demand is modeled either linearly, $\mu(p;\theta)=\theta_0-\theta_1 p$, or exponentially, $\mu(p;\theta)=\exp(\theta_0-\theta_1 p)$, where $\theta:=(\theta_0,\theta_1)$. The noise $\boldsymbol{\zeta}$ follows a mean-zero normal distribution truncated to $[-100,100]$ with standard deviation $\sigma\in\{15,30,60\}$. Feasible prices are $\pSet:=\{(100-q)/100\times P: q\in\Q\}$, where $P$ is the product’s full price, $q\in \Q$ shows discount levels with $\Q:=\{0,15,30,45,60\}$. Instances with linear and exponential demand models use $P=10$ and $P=30$, respectively.
    
    \begin{figure}[t]
    	\centering
    	\caption{Revenue curves of NI, SI, and MI instances with linear (left) and exponential (right) mean demands.\looseness=-1}
    	\includegraphics[width=1\linewidth]{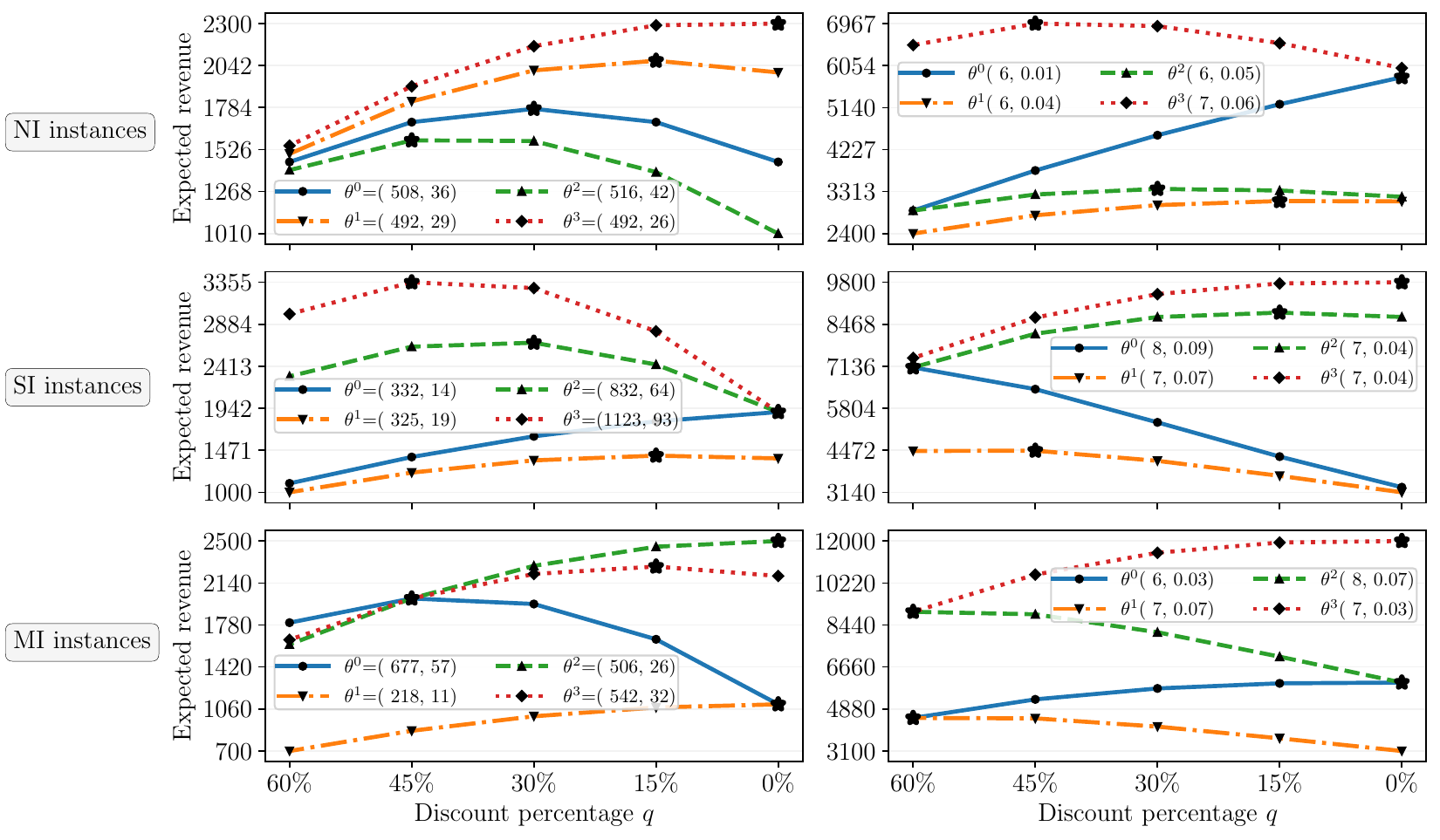}
    	\label{fig:revenue-curve}\vspace{-35pt}
    \end{figure}
    
    To construct the candidate set $\Theta$, we consider three classes of instances: non-intersecting (NI), single-price intersecting (SI), and multi-price intersecting (MI). These instances differ in how the mean demand functions $\{\mu(p;\theta) : \theta \in \Theta\}$ intersect across different prices $p \in \pSet$. Recall Assumption \ref{asm:separability}, which requires $\mu(p;\theta)\neq\mu(p;\theta')$ for every distinct pair $\theta\neq\theta'$ and every price $p\in\pSet$. NI instances satisfy this separability condition. SI instances violate it only at the CI price $\pCI$. MI instances violate separability at multiple prices, including $\pCI$. Figure \ref{fig:revenue-curve} illustrates these three designs for both linear and exponential demand. It contains six subplots: the left column corresponds to the linear model and the right column the exponential model, with the first, second, and third rows correspond to the NI, SI, and MI instances, respectively. Each subplot depicts the expected revenue $r(\cdotp;\theta):\pSet\mapsto[0,\infty)$ for the four candidate parameters $\theta\in\Theta$. The solid curve represents the true demand model $\theta^0$, and the dashed, dotted, and dash–dotted curves show the remaining candidates. The star marker on each curve with parameter $\theta$ shows the revenue‐maximizing price $p^\star(\theta):=\argmax\{r(p;\theta):p\in\pSet\}$.\looseness=-1

    \begin{figure}[t]
    	\centering
    	\caption{Customer arrival patterns for $\beta = 1.5$ (left), $\beta = -1.5$ (left-center), $\beta = 2$ (right-center), and $\beta = -2$ (right).}
    	\includegraphics[width=1\linewidth]{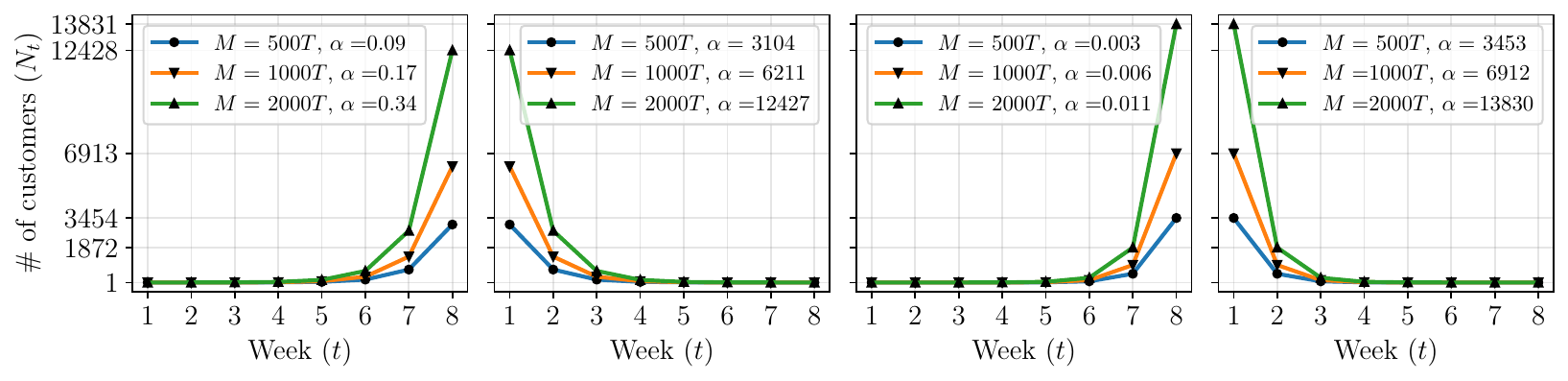}
    	\label{fig:numerics_arrival_patterns}\vspace{-45pt}
    \end{figure} 
    
    We model customer arrival patterns $(N_1, N_2, \ldots, N_T)$ using flat, exponentially increasing, and exponentially decreasing patterns by setting $N_t \coloneqq \lceil \alpha \exp(\beta (t-1)) \rceil$, where $\alpha$ controls the scale and $\beta$ the growth rate. We examine five values of $\beta$ to capture stable, delayed-hit, and early-hit sales patterns (top panel of Figure \ref{fig:pattern_DAS_evolution}): a flat pattern with $\beta = 0$, increasing patterns with $\beta \in \{1.5, 2\}$, and decreasing patterns with $\beta \in \{-1.5, -2\}$. To choose meaningful values of $\alpha$, we require the total customer count $\sum_{t=1}^{T} N_t$ to equal a fixed value $M$ under each pattern. Specifically, we set the horizon set to $T = 8$ and find $\alpha$ to satisfy $\sum_{t=1}^{T} \lceil \alpha \exp(\beta (t-1)) \rceil = M$ for each $M \in \{500T, 1000T, 2000T\}$. These choices reflect selling over eight weeks and observing 500, 1,000, or 2,000 customers per week under the flat pattern. Figure \ref{fig:numerics_arrival_patterns} depicts the resulting customer arrival patterns for each $\beta \in \{1.5, -1.5, 2, -2\}$ and $M \in \{500T, 1000T, 2000T\}$, and also reports the corresponding value of $\alpha$ for each $(\beta, M)$ pair. Flat patterns ($\beta = 0$) are omitted, as they would appear as horizontal lines with $\alpha = M/T$.\looseness=-1

\subsection{Policy Implementation}\label{sec:implementation}

    We implement the ARL policy by following the steps in Algorithm \ref{alg:ARL}. The algorithm requires specifying a probability level $\delta$ and a risk-adjusted revenue function $\eta(\cdot)$. To simplify analysis, we fix $\delta = 0.1$ and set $\eta(\cdot)$ to the $\alpha$-VaR risk-adjusted revenue from \S\ref{sec:risk-adjust-rev} with $\alpha = 0$, which returns the worst-case revenue in $\R(p; \A) = \{r(p; \theta) : \theta \in \A\}$. We compute the problem-specific quantities $n_\delta(p)$ and $c(p)$ in Algorithm \ref{alg:ARL} using their definitions in \S\ref{sec:alg}. For each instance and price $p \in \pSet$, we compute $c(p) = \min\big\{ |\mu(p; \theta) - \mu(p; \theta')| > 0 : (\theta, \theta') \in \Theta^2 \big\}$ by evaluating the gap $|\mu(p; \theta) - \mu(p; \theta')|$ using the formulation of the linear and exponential mean demand $\mu(p; \theta)$ for every pair $(\theta, \theta')$ in the candidate set $\Theta$, with the corresponding values visualized in each subplot of Figure \ref{fig:revenue-curve}. We then substitute $c(p)$ into the definition of $n_\delta(p)$, which also depends on the sub-exponential parameters $\nu$ and $b$ defined in Assumption \ref{assump:LightTailedDemand}. Since $\boldsymbol{\zeta}$ follows a truncated normal distribution, it is straightforward to verify that $\nu = 100$ and $b = 0$. Moreover, we implement the NRM and FTL policies. For the NRM policy, we use the $\alpha$-VaR risk-adjusted revenue with $\alpha = 0$ to enable a fair comparison with the ARL policy. For the FTL policy in Algorithm \ref{alg:FTL}, we set $\delta = 0.1$, consistent with the ARL policy, and draw the initial estimate $\thetaFTL{1}$ uniformly from $\Theta$ to reflect the absence of prior information about the true demand parameter $\theta^0$.\looseness=-1
    
    We simulate each policy over $5,000$ demand sample paths, chosen to ensure the standard error of all revenue estimates remains within $1\%$ of their means. We compare policy performance using two metrics: optimality gap and relative VaR (RVaR). The optimality gap for a policy is defined as the average over $5,000$ sample paths of the policy's total revenue over $T = 8$ weeks minus the expected total revenue under the CI policy, i.e., $Mr(\pCI; \theta^0)$, expressed as a percentage of the latter. To interpret the optimality gap, assume its value for a policy is $g\%$. This means the policy's expected total revenue is $g\%$ worse than that of the CI policy. The RVaR is defined as the 95\% value at risk of the total revenue over $5,000$ sample paths minus the expected total revenue under the CI policy, expressed as a percentage of the latter. To interpret the RVaR, assume its value for a policy is $\nu\%$. Then, there is at least a 5\% chance that the realized cumulative revenue from the policy is $\nu\%$ worse than the expected total revenue under the CI policy. Thus, pricing strategies with low expected gap and RVaR are preferable, as they achieve both high expected revenue and low downside risk.\looseness=-1

    \begin{figure}[h]
    	\centering
    	\caption{Distributions of optimality gap (left) and RVaR (right) across all instances.}
    	\includegraphics[width=1\linewidth]{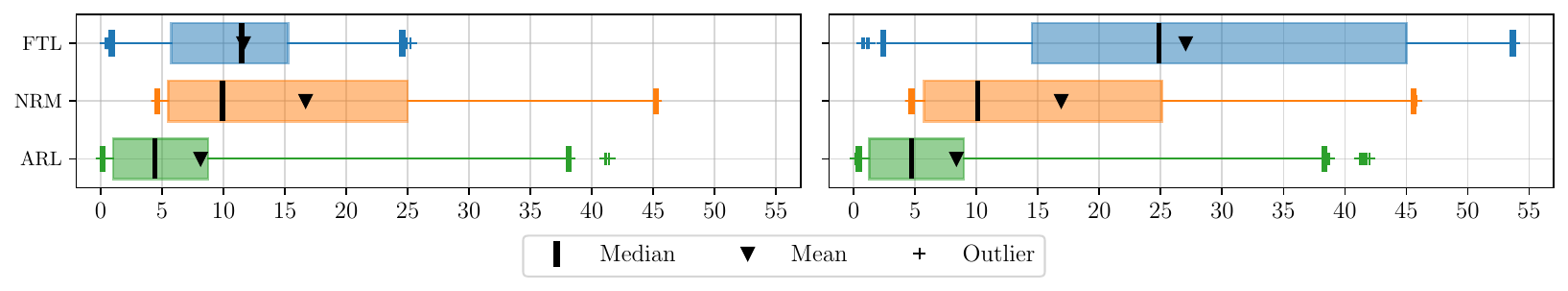}
    	\label{fig:gap_rvar_overall}\vspace{-50pt}
    \end{figure} 
\subsection{Results and Insights}\label{sec:results}

    Figure \ref{fig:gap_rvar_overall} displays the distributions of optimality gap and RVaR percentages for the ARL, NRM, and FTL policies across all test instances. The left panel reports optimality gaps, while the right panel presents RVaRs. In each panel, three horizontally aligned boxplots labeled ARL, NRM, and FTL summarize the performance of each policy. The median is represented by a horizontal bar, the mean as a triangle, and outliers as isolated plus signs. The ARL, NRM, and FTL policies achieve mean optimality gaps of 8\%, 16\%, and 12\%, respectively, and mean RVaRs of 8\%, 17\%, and 26\%. Thus, on average across instances, the ARL policy achieves 8\% higher mean revenue than the NRM policy and 4\% higher mean revenue than the FTL policy. In terms of downside risk, the 5\% tail revenue of the ARL policy outperforms that of the NRM and FTL policies by 9\% and 18\%, respectively. These results underscore the substantial value of the adaptation mechanism in ARL. Specifically, they demonstrate that ARL's smooth transition from risk mitigation to regret minimization yields stronger performance in both mean revenue and downside risk compared to the NRM and FTL policies, which focus solely on risk and regret, respectively. 

    \begin{figure}[h]
    	\centering
    	\caption{Distributions of optimality gap (left) and RVaR (right) across different customer arrival patterns.}
    	\includegraphics[width=1\linewidth]{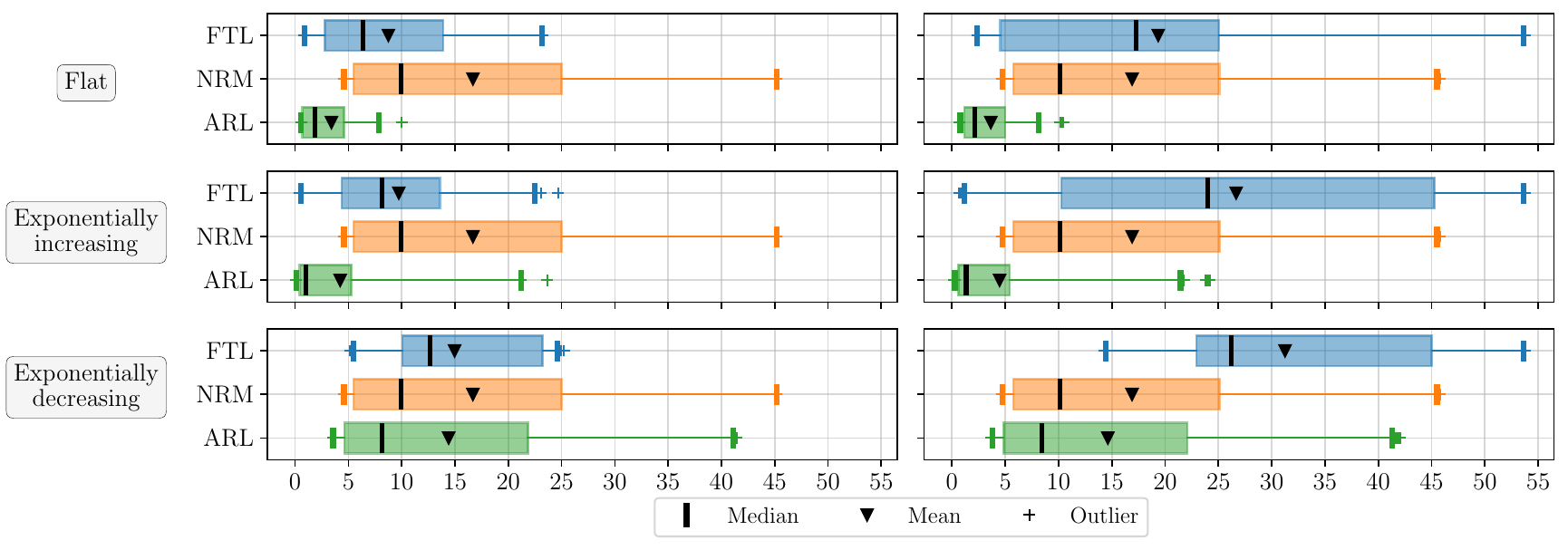}
    	\label{fig:gap_rvar_pattern}\vspace{-45pt}
    \end{figure} 
    
    Figure \ref{fig:gap_rvar_pattern} breaks down the distributions of optimality gap and RVaR for the ARL, NRM, and FTL policies across three customer arrival patterns: flat, exponentially increasing, and exponentially decreasing. Under the flat arrival pattern, where the same number of customers arrive each week, the ARL policy achieves the lowest mean optimality gap at 4\%, followed by the FTL policy at 9\%, and the NRM policy at 16\%. A similar trend holds for RVaR: the ARL policy maintains a mean RVaR of 4\%, while the FTL and NRM policies exhibit higher mean RVaRs of around 18\%. Under the exponentially increasing pattern, where most customers arrive during the final weeks, the ARL policy remains the best performer with a mean optimality gap of 5\%, followed by the FTL and NRM policies with mean gaps of approximately 10\% and 17\%, respectively. The corresponding mean RVaRs are 5\% for the ARL policy, 26\% for the FTL policy, and 16\% for the NRM policy. In the exponentially decreasing setting, where most customer arrive during the initial weeks, all three policies result in similar mean optimality gaps of around 15\%. However, the ARL and NRM policies maintain mean RVaRs close to 15\%, whereas the FTL policy results in the mean RVaR of 31\%. \looseness=-1

    Figure \ref{fig:gap_rvar_pattern} highlights the significant value of the ARL policy in adapting to different customer arrival patterns. Under the flat and exponentially increasing patterns, where early periods provide opportunities for learning and later periods can be used to optimize revenue, both the ARL and FTL policies achieve similar mean revenues and outperform the NRM policy, demonstrating the advantages of learning-focused methods. However, between the two, ARL substantially outperforms FTL in terms of downside risk. Specifically, the 5\% tail revenue of the ARL policy is, on average, 14\% higher than that of FTL under the flat pattern and 21\% higher under the exponentially increasing pattern. Under the exponentially decreasing pattern, where early pricing decisions are critical and there is limited opportunity to correct mistakes later, the 5\% tail revenues of the ARL and NRM policies are approximately 16\% higher than that of FTL on average, while all three policies yield comparable mean revenues. Together, these results underscore the power of adaptivity in ARL. That is, by adjusting to the unknown customer arrival pattern, ARL dynamically balances regret and risk, without the need to manually align these objectives. In contrast, benchmark policies that focus on only one of these objectives fall short under certain arrival patterns and evaluation metrics.

    \begin{table}[h]
    \renewcommand{\arraystretch}{.8}
        \centering
        \caption{Optimality gap and RVaR differences between ARL and FTL across NI, SI, and MI instances.}
        \resizebox{\textwidth}{!}{
            \begin{tabular}{l@{\hspace{1.5cm}}c@{\hspace{1.5cm}}c@{\hspace{1.5cm}}c}
            \hline
             & \textbf{NI Instances} & \textbf{SI Instances} & \textbf{MI Instances} \\
            \hline
            \textbf{Mean optimality gap difference, ARL$-$FTL (\%)}   & 1.65\%  & -13.21\% & -4.78\% \\
            \textbf{Mean RVaR difference, ARL$-$FTL (\%)}  & -1.34\% & -34.49\% & -24.28\% \\
            \hline
            \end{tabular}
        \label{tab:ARL-FTL-instance}
        }
    \end{table}\vspace{-15pt}
    Table \ref{tab:ARL-FTL-instance} reports the average differences in optimality gap and RVaR between the ARL and FTL policies across NI, SI, and MI instances, excluding those with exponentially decreasing arrival patterns because they offer limited opportunity to learn the true demand parameter. Negative percentages indicate ARL outperforms FTL, while positive values indicate the reverse. On NI instances, both policies achieve mean optimality gap and RVaR within 2\% of each other. On SI instances, ARL outperforms FTL by 13\% in mean optimality gap and 34\% in mean RVaR, and on MI instances, by 5\% and 24\%, respectively. These results highlight that ARL provides substantial improvements on SI and MI instances that do not satisfy Assumption \ref{asm:separability}. Thus, when separability fails, which is the expected case, ARL delivers meaningful gains over FTL in both average and downside performance, driven by its use of set estimation rather than point estimation, enhancing both learning (lower mean optimality gap) and risk mitigation (lower RVaR).\looseness=-1

\begin{figure}[t]
    	\centering
    	\caption{Illustrating the evolution of the DAS  under exponentially decreasing (left), flat (middle), and exponentially increasing (right) customer arrival patterns.}
    	\includegraphics[width=1\linewidth]{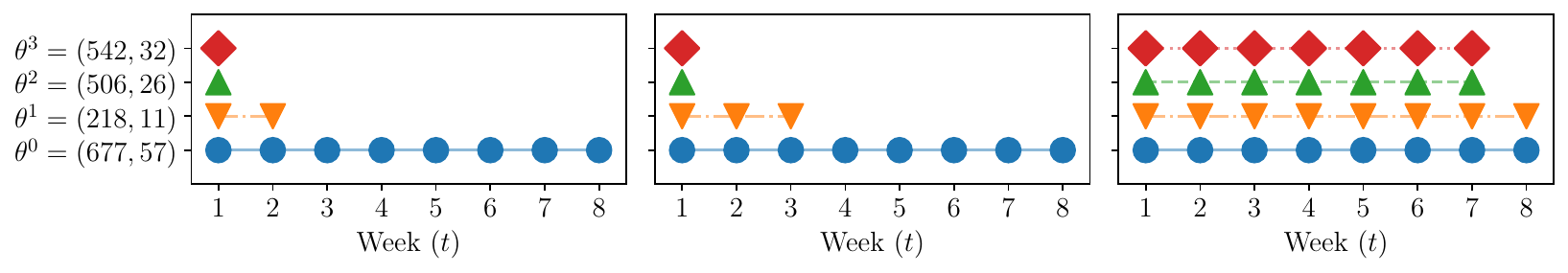}
    	\label{fig:das_evolution}\vspace{-40pt}
    \end{figure}

    Figure~\ref{fig:das_evolution} illustrates ARL's learning power using a representative MI instance shown in the bottom-left subplot of Figure~\ref{fig:revenue-curve}, with a demand noise standard deviation of $\sigma = 60$ and total store traffic of $M = 500T$. The left, middle, and right subplots correspond to exponentially decreasing, flat, and exponentially increasing arrival patterns, respectively. Each subplot displays weeks on the x-axis and the four candidate demand parameters $\{\theta^0, \theta^1, \theta^2, \theta^3\}$ on the y-axis. A solid marker at a given week indicates that the corresponding demand parameter is included in the DAS in more than at that time. The true demand parameter $\theta^0 = (677, 57)$ is marked with a circle, while the other parameters are represented using up-triangle, down-triangle, and diamond markers, consistent with Figure \ref{fig:revenue-curve}. For this MI instance, the true model coincides with the down-triangle model at the $q = 0\%$ discount level and with the up-triangle and diamond models at the $q =45\%$ discount level.\looseness=-1

    Under the exponentially decreasing pattern (left subplot), the ARL policy plays the 0\% discount level in the first week and collects enough data to eliminate $\theta^3$ and $\theta^2$ from the DAS, as their mean demands do not intersect with the true model. In the second week, ARL switches to the 15\% discount level, which maximizes the risk-adjusted revenue function given that the 0\% level is no longer partially informative, as the DAS only includes $\theta^0$ and $\theta^1$, which intersect at the 0\% level. By the end of the second week, sufficient data is collected to remove $\theta^1$, leaving the DAS with only the true model and resulting in a demand identification time of $t = 2$. A similar pattern occurs under the flat arrival pattern (middle subplot) but the demand identification time shifts to $t = 3$, indicating that one additional week is needed, compared to the exponentially decreasing pattern, to gather sufficient data at the 15\% discount level to eliminate $\theta^1$. Finally, under the exponentially increasing pattern (right subplot), the DAS retains all four candidate parameters until week $t = 7$ due to limited early data. Only in the final weeks does ARL collect enough observations to eliminate $\theta^3$ and $\theta^2$ from the DAS. Thus, the demand identification time is $t = 9$ under this arrival pattern, as insufficient data is available earlier in the season to confidently identify $\theta^0$ within the eight-week selling horizon. 
    
    Figure \ref{fig:das_evolution} illustrates the adaptive mechanics of the ARL policy. When the arrival pattern is favorable (left and middle subplots), ARL quickly identifies the true demand parameter and shifts early from risk mitigation to revenue maximization. In contrast, when the arrival pattern is unfavorable (right subplot), ARL retains all candidate models for most of the horizon, maintaining a cautious stance to mitigate risk .Only near the end, when sufficient data becomes available, does it reduce its level of risk aversion. Unlike ARL, the FTL policy fails to distinguish between $\theta^0$ and $\theta^1$ in this MI instance. These observations highlight ARL's ability to adapt its pace of model elimination to the timing and volume of observed data, accelerating learning when early information is abundant and prioritizing caution when learning opportunities are limited.
        \section{Conclusion}\label{sec:conclusion}
    Pricing under demand model ambiguity and unknown customer arrival patterns poses a significant challenge, especially when price updates are infrequent. With limited opportunities to adjust prices, a suboptimal price based on an inaccurate demand estimate can result in a substantial revenue loss. We tackle this challenge by developing an Adaptive Risk Learning (ARL) framework that learns the true demand parameters from online demand data and mitigates the risk of revenue loss due to model ambiguity and unknown arrival patterns.\looseness=-1

    ARL maintains a data-driven ambiguity set (DAS) that evolves over time by eliminating implausible demand parameters using no-regret logic. This set acts as a real-time indicator of model uncertainty: a large DAS reflects high ambiguity, while a smaller one signals greater confidence. The ARL policy adapts accordingly by prioritizing risk mitigation when the DAS is large and gradually shifting to regret minimization as the DAS shrinks. This adaptive behavior enables ARL to bridge the gap between the non-adaptive risk-mitigating (NRM) policy and the regret-focused follow-the-leader (FTL) policy. We establish the stochastic convergence of the DAS, derive the ARL policy's regret bound that explicitly depends on the customer arrival pattern, and show that ARL outperforms both NRM and FTL in terms of revenue assessments and regret. Our numerical results further validate these findings: ARL consistently achieves higher average revenue and significantly reduces downside risk across a broad range of demand models and arrival patterns. \looseness=-1
    
    More broadly, our methodology provides an intuitive, ambiguity-set-based extension of the well-known FTL approach, enabling the active balancing of risk and regret. This trade-off between regret and risk, together with the ``sticky-decision'' nature of our constrained dynamic pricing problem, also arises in other contexts, such as demand response in energy, ad allocation with batched updates, and omnichannel retailing. A natural direction for future work is to extend ARL to non-stationary demand environments.
	    {
	    	\OneAndAHalfSpacedXI
	        \bibliographystyle{informs2014} 
	        \bibliography{references}

@article{Feng_How_2018,
author = {Feng, Qi and Shanthikumar, J. George},
title = {How Research in Production and Operations Management May Evolve in the Era of Big Data},
journal = {Production and Operations Management},
volume = {27},
number = {9},
pages = {1670-1684},
year = {2018}
}

@article{besbes2011minimax,
  title={On the minimax complexity of pricing in a changing environment},
  author={Besbes, Omar and Zeevi, Assaf},
  journal={Operations research},
  volume={59},
  number={1},
  pages={66--79},
  year={2011},
  publisher={INFORMS}
}

@article{keskin2025nonstationary,
	title={The nonstationary newsvendor: Data-driven nonparametric learning},
	author={Keskin, N Bora and Min, Xu and Song, Jing-Sheng Jeannette},
	journal={Available at SSRN 3866171},
	year={2025},
	volume={},
	number={},
	pages={},
}

@article{gong2024bandits,
  title={Bandits atop reinforcement learning: Tackling online inventory models with cyclic demands},
  author={Gong, Xiao-Yue and Simchi-Levi, David},
  journal={Management Science},
  volume={70},
  number={9},
  pages={6139--6157},
  year={2024},
  publisher={INFORMS}
}

@article{chen2021data,
  title={Data-driven inventory control with shifting demand},
  author={Chen, Boxiao},
  journal={Production and Operations Management},
  volume={30},
  number={5},
  pages={1365--1385},
  year={2021},
  publisher={SAGE Publications Sage CA: Los Angeles, CA}
}

@article{keskin2017chasing,
  title={Chasing demand: Learning and earning in a changing environment},
  author={Keskin, N Bora and Zeevi, Assaf},
  journal={Mathematics of Operations Research},
  volume={42},
  number={2},
  pages={277--307},
  year={2017},
  publisher={INFORMS}
}

@article{besbes2015non,
  title={Non-stationary stochastic optimization},
  author={Besbes, Omar and Gur, Yonatan and Zeevi, Assaf},
  journal={Operations research},
  volume={63},
  number={5},
  pages={1227--1244},
  year={2015},
  publisher={INFORMS}
}

@techreport{mckinsey_dynamic_pricing_chemicals,
  title   = {Dynamic pricing: Using digital and analytics to take value pricing in the chemical industry to the next level},
  author  = {{McKinsey}},
  institution = {McKinsey \& Company},
  year    = {2019},
  urldate = {2025-10-02}
}

@techreport{McKinsey_2016,
	title = {Omnichannel compendium for retailers},
	author = {{McKinsey}},
    year    = {2019},
    institution = {McKinsey \& Company},
    urldate = {2025-10-02}
}

@article{den2019dynamic,
  title={Dynamic pricing with demand learning and reference effects},
  author={den Boer, Arnoud V and Keskin, N Bora},
  journal={Management Science},
  volume={68},
  number={10},
  pages={7112--7130},
  year={2022},
  publisher={INFORMS}
}

@techreport{WE_2020,
	title = {The future of unified commerce},
	author = {{Windstream Enterprise}},
	year = {2020}
}

@book{foss2011introduction,
  title={An introduction to heavy-tailed and subexponential distributions},
  author={Foss, Sergey and Korshunov, Dmitry and Zachary, Stan and others},
  volume={6},
  year={2011},
  publisher={Springer}
}

@article{kalai2005efficient,
  title={Efficient algorithms for online decision problems},
  author={Kalai, Adam and Vempala, Santosh},
  journal={Journal of Computer and System Sciences},
  volume={71},
  number={3},
  pages={291--307},
  year={2005},
  publisher={Elsevier}
}

@article{wainwright2015basic,
	title={Course on Mathematical Statistics, chapter 2: Basic tail and concentration bounds},
	author={Wainwright, Martin},
	journal={University of California at Berkeley, Department of Statistics},
	year={2015}
}

@incollection{blum1998line,
  title={On-line algorithms in machine learning},
  author={Blum, Avrim},
  booktitle={Online algorithms},
  pages={306--325},
  year={1998},
  publisher={Springer}
}

@article{chen2015recent,
  title={Recent developments in dynamic pricing research: multiple products, competition, and limited demand information},
  author={Chen, Ming and Chen, Zhi-Long},
  journal={Production and Operations Management},
  volume={24},
  number={5},
  pages={704--731},
  year={2015},
  publisher={Wiley Online Library}
}

@incollection{araman2010revenue,
  title        = {Revenue Management with Incomplete Demand Information},
  author       = {Araman, Victor F. and Caldentey, Ren{\'e}},
  booktitle    = {Wiley Encyclopedia of Operations Research and Management Science},
  year         = {2010},
  publisher    = {Wiley}
}

@article{den2015dynamic,
  title={Dynamic pricing and learning: historical origins, current research, and new directions},
  author={den Boer, Arnoud V},
  journal={Surveys in operations research and management science},
  volume={20},
  number={1},
  pages={1--18},
  year={2015},
  publisher={Elsevier}
}

@article{bitran2003overview,
  title={An overview of pricing models for revenue management},
  author={Bitran, Gabriel and Caldentey, Ren{\'e}},
  journal={Manufacturing \& Service Operations Management},
  volume={5},
  number={3},
  pages={203--229},
  year={2003},
  publisher={INFORMS}
}

@article{elmaghraby2003dynamic,
  title={Dynamic pricing in the presence of inventory considerations: Research overview, current practices, and future directions},
  author={Elmaghraby, Wedad and Keskinocak, P{\i}nar},
  journal={Management science},
  volume={49},
  number={10},
  pages={1287--1309},
  year={2003},
  publisher={INFORMS}
}

@article{pakiman2020self,
  title={Self-guided approximate linear programs: randomized multi-shot approximation of discounted cost markov decision processes},
  author={Pakiman, Parshan and Nadarajah, Selvaprabu and Soheili, Negar and Lin, Qihang},
  journal={Management science},
  volume={71},
  number={4},
  pages={3384--3404},
  year={2025},
  publisher={INFORMS}
}

@inproceedings{shah2020reinforcement,
  title={On reinforcement learning for turn-based zero-sum Markov games},
  author={Shah, Devavrat and Somani, Varun and Xie, Qiaomin and Xu, Zhi},
  booktitle={Proceedings of the 2020 ACM-IMS on Foundations of Data Science Conference},
  pages={139--148},
  year={2020}
}

@article{BZ2009,
author = {O. Besbes and A. Zeevi},
title = {Dynamic Pricing Without Knowing the Demand Function: Risk Bounds and Near-Optimal Algorithms},
journal = {Operations Research},
volume = {57},
number = {6},
pages = {1407-1420},
year = {2009},
}

@article{BZ2012,
author = {O. Besbes and A. Zeevi},
title = {Blind Network Revenue Management},
journal = {Operations Research},
volume = {60},
number = {6},
pages = {1537-1550},
year = {2012},
}

@article{BZ2015,
 author = {Besbes, O. and Zeevi, A.},
 title = {On the Surprising Sufficiency of Linear Models for Dynamic Pricing with Demand Learning},
 journal = {Management Science},
 issue_date = {April 2015},
 volume = {61},
 number = {4},
 month = apr,
 year = {2015},
 pages = {723--739},
 numpages = {17},
 }

@article{WDY2014,
author = {Z. Wang and S. Deng and Y. Ye},
title = {Close the Gaps: A Learning-While-Doing Algorithm for Single-Product Revenue Management Problems},
journal = {Operations Research},
volume = {62},
number = {2},
pages = {318--331},
year = {2014},
}

@article{CJD2018,
author = "Chen, Q and Jasin, S and Duenyas, I",
title = "A nonparametric self-adjusting control for joint learning and optimization of multi-product pricing with finite resource capacity",
journal={Mathematics of Operations Research},
  volume={44},
  number={2},
  pages={601--631},
  year={2019}
}

@unpublished{LJS2019,
author = "Y. Lei and S. Jasin and A. Sinha",
title = "Near-Optimal Bisection Search for Nonparametric Dynamic Pricing with Inventory Constraint",
note = "Working paper, University of Michigan, Ann Arbor, MI",
year=2019
}

@article{CCA2019,
author = "Chen, Boxiao and Chao, Xiuli and Ahn, Hyun-Soo",
title = "Coordinating pricing and inventory replenishment with nonparametric demand learning",
journal={Operations Research},
  volume={67},
  number={4},
  pages={1035-1052},
  year={2019}
}

@article{CCS2019,
author = {Chen, Boxiao and Chao, Xiuli and Shi, Cong},
title = {Nonparametric Learning Algorithms for Joint Pricing and Inventory Control with Lost Sales and Censored Demand},
journal = {Mathematics of Operations Research},
volume = {46},
number = {2},
pages = {726-756},
year = {2021},
}

@article{chen2023robust,
  title={Robust dynamic pricing with demand learning in the presence of outlier customers},
  author={Chen, Xi and Wang, Yining},
  journal={Operations Research},
  volume={71},
  number={4},
  pages={1362--1386},
  year={2023},
  publisher={INFORMS}
}

@article{perakis2024dynamic,
  title={Dynamic pricing with unknown nonparametric demand and limited price changes},
  author={Perakis, Georgia and Singhvi, Divya},
  journal={Operations Research},
  volume={72},
  number={6},
  pages={2726--2744},
  year={2024},
  publisher={INFORMS}
}

@phdthesis{broder2011online,
  title={Online algorithms for revenue management},
  author={Broder, Josef},
  school={Cornell University},
  year={2011},
}

@article{keskin2014dynamic,
  title={Dynamic pricing with an unknown demand model: Asymptotically optimal semi-myopic policies},
  author={Keskin, N Bora and Zeevi, Assaf},
  journal={Operations Research},
  volume={62},
  number={5},
  pages={1142--1167},
  year={2014},
  publisher={INFORMS}
}

@article{CSW2017,
  title={Dynamic pricing and demand learning with limited price experimentation},
  author={Cheung, Wang Chi and Simchi-Levi, David and Wang, He},
  journal={Operations Research},
  volume={65},
  number={6},
  pages={1722--1731},
  year={2017},
  publisher={INFORMS}
}

@article{shalev2011online,
  title={Online learning and online convex optimization},
  author={Shalev-Shwartz, Shai and others},
  journal={Foundations and trends in Machine Learning},
  volume={4},
  number={2},
  pages={107--194},
  year={2011}
}

@article{shin2023dynamic,
  title={Dynamic pricing with online reviews},
  author={Shin, Dongwook and Vaccari, Stefano and Zeevi, Assaf},
  journal={Management Science},
  volume={69},
  number={2},
  pages={824--845},
  year={2023},
  publisher={INFORMS}
}

@article{chen2020data,
  title={Data-based dynamic pricing and inventory control with censored demand and limited price changes},
  author={Chen, Boxiao and Chao, Xiuli and Wang, Yining},
  journal={Operations Research},
  volume={68},
  number={5},
  pages={1445--1456},
  year={2020},
  publisher={INFORMS}
}
	    }
	    \newpage
	     
\ECSwitch
\ECHead{
    \centering
    Electronic Companion to \\[20pt]
	\theRUNTITLE
}\vspace{1cm}

\section{Proofs}
\begin{proposition}\label{prop:PI}
    The set of partially informative prices $\PIP(\A)$ with respect to any nonempty ambiguity set $\A \subseteq \Theta$ is nonempty.
\end{proposition}
\proof{\textbf{Proof of Proposition \ref*{prop:PI}.}}%
    Fix an ambiguity set $\A \subseteq \Theta$. Because $\A$ is nonempty, we have $|\A|\ge 1$. If $|\A| = 1$, then $\PIP(\A)$ is nonempty by definition in \S \ref*{sec:partially-inform-price}, which sets $\PIP(\A) = \pSet$. If $|\A| \geq 2$, then there exist two distinct models with parameters $\theta, \theta' \in \A \subseteq \Theta$. Assumption \ref*{asm:deman-model-set} ensures the existence of a price $p \in \pSet$ such that $|\mu(p; \theta) - \mu(p; \theta')| > 0$. Therefore, $p$ is a partially informative price with respect to $\A$, and $p \in \PIP(\A)$, implying $\PIP(\A) \ne \varnothing$ is nonempty.
\hfill$\Box$\endproof

\begin{theorem}[Two-sided sub-exponential tail bound]\label{thm:Azuma}
    Let $X$ be a mean-zero sub-exponential random variable with parameters $(u,w)$. Then, for every $\zeta\ge 0$, we have the following concentration inequality:
    \[
    \mathbb{P}\left(\left|X\right| \ge \zeta \right)  \ \le \ 
    2\exp\left(-\min\left\{ \frac{\zeta^2}{ 2w^2},\   \frac{\zeta}{2u} \right\}\right)
    \]
	% Let $\{H_k\}_{k=1}^{\infty}$ be a martingale difference sequence with respect to filtration $\{\mathcal{H}_k\}_{k=1}^{\infty}$. Suppose that there are constants $u,w\ge0$ such that the following inequality holds almost surely:  
 %    \[
 %        \mathbb{E}_{H_k}[\exp(\lambda H_k) \ | \ \mathcal{H}_{k-1}] \le \exp(\frac{\lambda^2 w^2}{2}), \quad \forall \lambda\in(-{1}/{u},{1}/{u}), \ \forall k\ge1.
 %    \]
 %    Then, for every $\zeta\ge 0$, we have the following concentration inequality:
	% \[\mathbb{P}\left(\left|\sum_{k=1}^{K} H_k \right| \ge \zeta \right) \ \le \ 2\exp\left(-\min\left\{ \frac{\zeta^2}{ 2w^2K},\   \frac{\zeta}{2u} \right\}\right) .\]
\end{theorem} 
\proof{\textbf{Proof of Theorem \ref*{thm:Azuma}.}}%
    Please refer to Proposition 2.2 in \citealt{wainwright2015basic}.
    % Please refer to Theorem 2.3  (b) in \citealt{wainwright2015basic}.
\hfill$\Box$\endproof

\proof{\textbf{Proof of Theorem \ref*{thm:DAS-evolution}.}}%
    Fix the customer arrival pattern $(N_1, \dots, N_T)$, $\delta \in (0,1]$, and an iteration $t\in\indexSet{T}$. Define the abbreviations $\hat M_t := |\D(\hat p_t)|$, $\hat{n}_t := n_\delta(\hat{p}_{t})$, $\hat c_t := c(\hat p_t)$, and $\psi(c):=\min \left \{{c^2}/{8v^2}, {c}/{4b} \right\}$. For every sequences $(\hat\A_{1}, \dots, \hat\A_{T})$ and $(\hat{p}_{1}, \dots, \hat{p}_{T})$ obtained from Algorithm \ref*{alg:ARL}, we show that the DAS update rule in (\ref*{eq:update_U}) and the inclusion property $\theta^0 \in \hat{\A}_{t}$ hold with probability at least $1 - \delta$. 
    At time $t$, if $\hat M_t < \hat{n}_t$, then the identity $\hat{\A}_2 = \hat{\A}_1$ holds with probability one by the if-condition in Algorithm~\ref*{alg:ARL}. This directly implies that the update in (\ref*{eq:update_U}), which states that $\hat{\A}_{t+1} = \hat{\A}_t$ holds with probability at least $1 - \delta$ when $\hat M_t < \hat{n}_t$, is satisfied. Now, if $\hat M_t \ge \hat{n}_t$, then the else-condition in Algorithm \ref*{alg:ARL} implies $\hat\A_{t+1} = \big\{\theta  \in \hat\A_{t} \big|   \ell\big(\theta ;\D(\hat p_t)\big) <  \hat{c}_t /2 \big\}$. We next show that the DAS $\hat\A_{t+1}$ satisfies $\hat\A_{t+1} = \hat\A_{t}\cap \I(\hat p_t)$ with probability $1-\delta$. For each $i\in\indexSet{\hat M_t}$, define the following random variable:
    \[
        z_{ti}:= d_i(\hat p_t) - \mu(\hat p_t; \theta^0).
    \]
    Given $\hat\A_{t}$, the price $\hat{p}_t = \argmax\{\eta(p;\hat\A_t): p\in  \PIP(\hat\A_t)\}$ and the random variable $\hat M_t= |\D(\hat p_t)|$ become deterministic, and the only stochastic element in $z_{ti}$ is the random variable $d_i(\hat p_t)$. Because for every price $p\in \mathcal{P}$, the demand distribution $\distD(\cdot)$ is sub-exponential with parameters $(v, b)$ due to Assumption \ref*{assump:LightTailedDemand}, each i.i.d. random variable $d_i(\hat p_t)$ is sub-exponential with the same parameters. It can be also seen that the mean-zero random variable $z_{ti}|\hat\A_{t}$ is also sub-exponential with parameters $(v, b)$ for each $i\in\indexSet{\hat M_t}$. That is, for each $\lambda\in(-{1}/{b},{1}/{b})$ and $i\ge1$, we have\looseness=-1 
    %Thus, $\{z_{ti} | \hat\A_{t}\}_{i=1}^\infty$ is a martingale difference sequence, as $\expt[z_t|\hat\A_{t}]=0$ and $\expt[|z_t|]<\infty$. 
    \begin{equation*} \label{subexponential}
        \expt\big[\exp\left(\lambda z_{ti}\right) \ \big | \ \hat \A_{t}  \big] \ \le \  \exp(\frac{\lambda^2 v^2}{2}).
    \end{equation*} 
    It is also easy to see that $z_t:=\sum_{i=1}^{\hat M_t}z_{ti}$, which is the sum of $\hat M_t$ sub-exponential random variables $z_{ti}$, is itself sub-exponential with parameters $(v\sqrt{\hat M_t}, b)$. Therefore, applying the sub-exponential tail bound in Theorem \ref*{thm:Azuma} to the random variable $z_t | \hat \A_{t}$, results in the following inequality for every $\zeta>0$:
    \begin{eqnarray}
        \mathbb{P}\left(\left|z_t \right| \ge \zeta  \ \big | \ \hat \A_{t} \right) \ \le \  2\exp\left(-\min \left \{\frac{\zeta^2}{2v^2 \hat M_t}, \frac{\zeta}{2b} \right\}\right), \nonumber
    \end{eqnarray}
    By setting $\zeta:= \hat c_t \hat M_t/2$, we obtain the following tail bound:
    \begin{eqnarray}
        \mathbb{P}\left(\frac{\left|z_t  \right|}{\hat M_t} < \frac{\hat c_t}{2} \ \bigg | \ \hat \A_{t}\right) \ \ge \ 1 - 2\exp\left(-\hat M_t   \min \left \{\frac{\hat{c}_t^2}{8v^2}, \frac{\hat{c}_t}{4b} \right\}\right)= 1 - 2\exp(-\hat M_t\psi(\hat{c}_t)). \nonumber
    \end{eqnarray}
    Combining the inequality $M_{t} \ge \bar{n}_{t} = n(\hat p_t)$ with the definition of $\psi(\cdot)$ at the beginning of this proof and $\bar{n}_{t} = n(\hat p_t) = {\ln(2/\delta)}/{\psi(\hat{c}_t)}$ in \S \ref*{sec:alg}, we obtain $\hat M_t\psi(\hat{c}_t) \ge \bar{n}_{t}\psi(\hat{c}_t)=\ln(2/\delta)$, which, together with the above tail bound, results in the following inequality:
    \begin{equation}\label{eq:Z-1}
        \mathbb{P}\left( \left|\frac{1}{\hat M_t}\sum_{i=1}^{\hat M_t}\left[d_i(\hat p_t)   -  \mu(\hat p_t; \theta^0) \right]\right| <  \frac{\hat{c}_t}{2} \ \Bigg | \ \hat \A_{t} \right) 
        =  \mathbb{P}\left(\frac{\left|z_t  \right|}{\hat M_t}  <  \frac{\hat{c}_t}{2} \ \bigg | \ \hat \A_{t}  \right)
        \ge  1 - 2\exp(- \bar{n}_{t} \psi(\hat{c}_t)) =  1-\delta.
    \end{equation}
    From  (\ref*{eq:Z-1}), for $\theta \in \Theta \backslash \I(\hat p_t)$, the following inequalities hold with probability of $1-\delta$:
    \begin{equation}\label{eq:mean-gap}
    \begin{aligned}
        \ell\big(\theta ;\D(\hat p_t)\big) 
            & = \left|\mu(\hat p_t; \theta) \ - \ \frac{1}{\hat M_t}\sum_{i=1}^{\hat M_t} d_i(\hat p_t )   \right|\\
            & =  \left|\Big(\mu(\hat p_t; \theta) - \mu(\hat p_t; \theta^0)\Big) \ - \ \frac{1}{\hat M_t}\sum_{i=1}^{\hat M_t} \Big(d_i(\hat p_t )-\mu(\hat p_t; \theta^0)\Big)   \right|\\
            & \ge \left|  \left|(\mu(\hat p_t; \theta) - \mu(\hat p_t; \theta^0))\right| \ - \ \left|\frac{1}{\hat M_t}\sum_{i=1}^{\hat M_t} (d_i(\hat p_t )-\mu(\hat p_t; \theta^0))\right| \right|\\
            & \ge \left|(\mu(\hat p_t; \theta) - \mu(\hat p_t; \theta^0))\right| \ - \  \left|\frac{1}{\hat M_t}\sum_{i=1}^{\hat M_t} (d_i(\hat p_t)-\mu(\hat p_t; \theta^0))\right| \\
            & \ge \hat{c}_t - \hat{c}_t/2
    \end{aligned}
    \end{equation}
    The first equality follows from the definition of loss function $\ell\big(\theta ;\D(\hat p_t)\big)$, noting that $\D(\hat p_t)$ includes demand data $d\in\D(\hat p_t)$ with $d = d_i(\hat p_t )$ for some $i\in\indexSet{\hat M_t}$. The second equality is obtained by adding and subtracting the term $\mu(\hat p_t; \theta^0)$. The first inequality follows from the reverse triangle inequality: $|a - b| \geq \big||a| - |b|\big|$ for real numbers $a$ and $b$. The second inequality holds due to $|a - b| \ge a - b$ for real numbers $a$ and $b$. The last inequality results from two key observations. First, the inequality $|\mu(\hat p_t; \theta) - \mu(\hat p_t; \theta^0)| \ge \hat{c}_t$ holds for every $\theta \in \Theta \backslash \I(\hat p_t)$. To see this, note that each model $\theta\notin \I(\hat p_t)$ does not intersect with $\theta^0$ at price $\hat{p}_1$, ensuring that gap $|\mu(\hat p_t; \theta) - \mu(\hat p_t; \theta^0)|$ is positive. From the definition of $\hat{c}_t$, this gap is at least $\hat{c}_t$. Hence we have $|\mu(\hat p_t; \theta) - \mu(\hat p_t; \theta^0)| \ge \hat{c}_t$ for every $\theta\notin \I(\hat p_t)$. Second, the negative of $\big|\big(\sum_{i=1}^{\hat M_t} (d_i(\hat p_t)-\mu(\hat p_t; \theta^0))\big)/\hat M_t\big|$ is lower bounded  by $(-\hat{c}_t/2)$ via (\ref*{eq:Z-1}) with probability $1 - \delta$.
    Next, we observe that for each $\theta \in \I(\hat p_t)$, the following inequalities hold with probability of $1-\delta$:
    \[
    \ell\big(\theta ;\D(\hat p_t)\big) 
            = \left|\mu(\hat p_t; \theta) \ - \ \frac{1}{\hat M_t}\sum_{i=1}^{\hat M_t} d_i(\hat p_t )   \right| 
            = \left|\mu(\hat p_t; \theta^0) \ - \ \frac{1}{\hat M_t}\sum_{i=1}^{\hat M_t} d_i(\hat p_t )   \right|
            \le \hat{c}_t/2.
    \]
    The leftmost equality follows directly from the definition of $\ell\big(\theta;\D(\hat p_t)\big)$. The second equality holds because $\mu(\hat p_t; \theta) = \mu(\hat p_t; \theta^0)$ for each $\theta \in \I(\hat p_t)$ by the definition of $\I(\hat p_t)$, and the inequality holds with probability at least $1 - \delta$ due to (\ref*{eq:Z-1}).
    Collectively, we showed that with probability $1 - \delta$, $\ell\big(\theta ;\D(\hat p_t)\big) \le \hat{c}_t/2$ for every $\theta \in \I(\hat p_t)$ and $\ell\big(\theta ;\D(\hat p_t)\big) \ge \hat{c}_t/2$ for every $\theta \in \Theta \backslash \I(\hat p_t)$. These two properties imply that the DAS $\hat{\A}_{t+1}$ satisfies the following identity with probability $1 - \delta$:
    \[
        \hat\A_{t+1}  = \hat\A_{t}\cap \left\{\theta  \in \Theta \ \Big| \  \ell\big(\theta ;\D(\hat p_t)\big) < \frac{c(\hat p_t)}{2}  \right\} = \hat\A_{t}\cap\I(\hat p_t),
    \]
    Note that the leftmost equality above follows from the update to $\hat\A_{t+1}$ from the else-condition of Algorithm \ref*{alg:ARL}. Therefore, update (\ref*{eq:update_U}) in Theorem \ref*{thm:DAS-evolution} is valid for $t=1$. Moreover, $\theta^0 \in \hat\A_{t+1}$ holds with probability $1 - \delta$ directly by combining the definition of $\ell(\theta^0;\D(\hat p_t))$ with (\ref*{eq:Z-1}), which results in:\looseness=-1
    \[
        \mathbb{P}\bigg( \ell(\theta^0;\D(\hat p_t)) <  \frac{\hat{c}_t}{2}\bigg) = \mathbb{P}\bigg( \bigg|\frac{1}{\hat M_t}\sum_{i=1}^{\hat M_t}\bigg[d_i(\hat p_t)   -  \mu(\hat p_t; \theta^0) \bigg]\bigg| <  \frac{\hat{c}_t}{2} \bigg) \ \ge \  1-\delta.
    \]
    Since the iteration $t\in\indexSet{T}$ was chosen arbitrarily, the above arguments show that, for every $t$, both (\ref*{eq:update_U}) and $\theta^0\in\hat\A_t$ hold with probability at least $1-\delta$. Thus, the proof is complete.
\hfill$\Box$\endproof

\proof{\textbf{Proof of Proposition \ref*{prop:stoch-to-det}.}}%
    \textbf{\underline{Part (1).}}
        In this part, we show that for every $k \in \indexSet{K}$, the identity $\hat{\A}_{t} = \bar{\A}_{k}$ for each $t \in \{\bar{t}_{k-1}(\delta)+1, \bar{t}_{k-1}(\delta)+2, \dots, \bar{t}_{k}(\delta)\}$ holds with probability at least $1 - (k-1)\delta$. We proceed by induction on $k$. Consider the following induction hypothesis for $k \in \indexSet{K}:$
        \begin{equation}\tag{IH$_k$}\label{eq:IH}
            \prob\big(\hat\A_t=\bar\A_k\big)\ge 1-(1-k)\delta, \quad  \forall t\in\big\{\bar t_{k-1}(\delta)+1,\dots,\bar t_k(\delta)\big\}. 
        \end{equation}
        Above, the probability operator $\prob(\cdot)$ is over the DAS $\hat\A_t$, which depends on ARL past pricing decisions and demand datapoints.\\
    \textbf{Base case.} 
        For $k=1$, we have $\bar\A_1 = \Theta$ from its definition in \S \ref*{subsec:ARL-theory}. Since Algorithm \ref*{alg:ARL} initializes the DAS to $\hat\A_1 = \Theta$, it follows that $\bar\A_1 = \hat\A_1 = \Theta$ and $\hat p_1 = \bar p_1 = \argmax\{\eta(p;\Theta) : p \in \PIP(\Theta)\}$ hold with probability one. In ARL, as long as the if-condition $|\D(\hat p_t)| \ge n_\delta(\hat p_t)$ is not satisfied, both the price $\hat p_1$ and the DAS $\hat\A_1 = \Theta$ remain unchanged. Therefore, we have $|\D(\hat p_t)| = |\D(\bar p_1)| = \sum_{s=1}^{t} N_s$. From the definition of $\bar t_1(\delta)$ in \S \ref*{subsec:ARL-theory}, time $t = \bar t_1(\delta)$ is the first period for which $\sum_{s=1}^{t} N_s \ge n_\delta(\bar p_1)$ holds. Thus, for every $t \in \{1,2,\dots,\bar t_{1}(\delta)\}$, we have $\hat\A_t = \bar\A_1 = \Theta$ with probability one, establishing (\ref*{eq:IH}) for the base case $k=1$.\\
        \textbf{Induction step.}
            % Fix step $\kappa \le K-1$. Assume \eqref{eq:IH} holds for every $k\le \kappa$. We show that \eqref{eq:IH} holds for $k=\kappa+1$. That is, for every $j=1,2,\dots,\kappa$, we have
            % \[
            %     \prob\left(\hat\A_t=\bar\A_j\right)\ge 1-(j-1)\delta, \quad  \forall t\in\big\{\bar t_{j-1}(\delta)+1,\dots,\bar t_j(\delta)\big\}. 
            % \]
            % Similarly, for the ARL price $\hat p_t$ optimized using $\hat \A_t$ and its deterministic analogue $\bar p_j$ optimized using $\bar \A_k$, where $\hat\A_t=\bar\A_j$ holds with probability  $1-(j-1)$, we have the following:
            % \[
            %     \prob\left(\hat p_t=\bar p_j\right)\ge 1-(j-1)\delta, \quad  \forall t\in\big\{\bar t_{j-1}(\delta)+1,\dots,\bar t_j(\delta)\big\}. 
            % \]
                % (ii) $\hat\A_t = \bar\A_2=\I(\bar p_1)$ and $\hat p_t=\bar p_2$ for every $t\in\big\{\bar t_1(\delta)+1,\dots,\bar t_2(\delta)\big\}$, and 
                % (iii) this continues to $k=\kappa$ for which $\hat\A_t = \bar\A_{\kappa}=\cap_{j=1}^{\kappa}\I(\bar p_{j})$ and $\hat p_t=\bar p_{\kappa}$ for every $t\in\big\{\bar t_{\kappa-1}(\delta)+1,\dots,\bar t_\kappa(\delta)\big\}$. 
        Fix step $\kappa \le K - 1$ and suppose (\ref*{eq:IH}) holds for $k = \kappa$. We show that it also holds for $k = \kappa + 1$. From the assumption in the induction step, we have:
        \[
            \prob\left(\hat{\A}_t = \bar{\A}_\kappa\right) \ge 1 - (\kappa - 1)\delta, \quad \forall t \in \left\{\bar{t}_{\kappa-1}(\delta)+1, \dots, \bar{t}_\kappa(\delta)\right\}.
        \]
        During this interval, Algorithm \ref*{alg:ARL} plays the fixed price $\hat{p}_t = \bar{p}_\kappa$ and updates the dataset $\D(\hat{p}_t)$, which satisfies $|\D(\hat{p}_t)| = |\D(\bar{p}_\kappa)| = \sum_{s = \bar{t}_{\kappa-1}(\delta)+1}^{t} N_s$. From the definition of $\bar{t}_\kappa(\delta)$, this is the first time the sample size $|\D(\hat{p}_t)|$ exceeds $n_\delta(\bar{p}_\kappa)$. Hence, at $t = \bar{t}_\kappa(\delta)$, the algorithm triggers an update to the DAS:
        \[
            \hat{\A}_{t+1} = \hat{\A}_t \cap \I(\hat{p}_t).
        \]
        We now verify the following identities hold with probability at least $1 - \kappa\delta$:
        \[
            \hat{\A}_{t+1} = \hat{\A}_t \cap \I(\hat{p}_t) = \bar{\A}_\kappa \cap \I(\bar{p}_\kappa) = \bar{\A}_{\kappa+1}.
        \]
        The leftmost equality holds with probability $1 - \delta$ by Theorem \ref*{thm:DAS-evolution} at $t = \bar{t}_\kappa(\delta)$, since the algorithm triggers an update to the DAS $t = \bar{t}_\kappa(\delta)$ (i.e., the else-condition in Algorithm \ref*{alg:ARL} occurs). The second equality is obtained via $\hat{\A}_t = \bar{\A}_\kappa$ that holds with probability $1 - (\kappa - 1)\delta$ by the inductive assumption, and $\hat{p}_t = \bar{p}_\kappa$ also holds with the same probability. The rightmost equality follows directly from the recursive definition of $\bar{\A}_{\kappa+1}$. Applying the union bound, we conclude that $\hat{\A}_{t+1} = \bar{\A}_{\kappa+1}$ holds with probability at least $(1- ((\kappa-1)\delta + \delta))$. In other words, we showed the following property:
        \[
            \prob\big(\hat\A_t=\bar\A_{\kappa+1}\big)\ge 1-\kappa\delta, \quad  t=t_\kappa(\delta)+1.
        \]
        At iteration $t = \bar{t}_\kappa(\delta) + 1$, the algorithm plays the price $\hat{p}_t = \bar{p}_{\kappa+1}$. It continues to play this price in every subsequent period $t \ge \bar{t}_\kappa(\delta) + 1$ until the number of customer arrivals at this price, $ |\D(\hat{p}_t)| = |\D(\bar{p}_{\kappa+1})| = \sum_{s = \bar{t}_\kappa(\delta) + 1}^{t} N_s,$ exceeds the threshold $n_\delta(\bar{p}_{\kappa+1})$. By definition, the earliest such time is $t = \bar{t}_{\kappa+1}(\delta)$. Therefore, for all $t \in \{\bar{t}_\kappa(\delta)+1, \dots, \bar{t}_{\kappa+1}(\delta)\}$, the DAS remains constant at $\hat{\A}_t = \bar{\A}_{\kappa+1}$ with probability at least $1 - \kappa\delta$. This completes the induction step and establishes that (\ref*{eq:IH}) holds for $k = \kappa + 1$.\\
        \textbf{Conclusion.} 
            The completed induction on $k$ establishes that (\ref*{eq:IH}) holds for every $k=1,2,\dots,K$. Thus, the proof of Part (1) is complete.\\
    \textbf{\underline{Part (2).}} 
        We now show that $\hat{\A}_{t} = \{\theta^0\}$ holds with probability $1-K\delta$ for all $t \in \{\tau(\delta)+1, \dots, T\}$, where $\tau(\delta):= \bar{t}_{K}(\delta)$. From Part (1) with $k=K$ and $t=\bar t_{K}(\delta)$, we have the identity $\hat\A_{\tau(\delta)}=\bar\A_K$, which holds with probability of $1-K\delta$. Recalling the definition of $\bar t_{K}(\delta)$ from \S \ref*{subsec:ARL-theory}, time $t = \bar t_{K}(\delta)$ is  the earliest period at which the ambiguity set $\hat{\A}_{1}=\bar{\A}_{1}=\Theta$ shrinks to $\bar{\A}_{K+1} = \{\theta^0\}$. From the definition of $\bar{\A}_{K+1}$, we have:
        \[
            \{\theta^0\} = \bar{\A}_{K+1} = \bar{\A}_{K}\cap \I(\bar p_K) = \hat{\A}_{\bar t_K(\delta)} \cap \I(\hat p_{\bar t_K(\delta)})  = \hat{\A}_{\bar t_K(\delta)+1}.
        \]
        The second equality from the left holds by the recursive definition of $\bar{\A}_{K+1}$ in \S \ref*{subsec:ARL-theory}. The third equality from the left follows from the identities $\hat\A_{\tau(\delta)} = \bar\A_K$ and $\tau(\delta) = \bar{t}_{K}(\delta)$, which together hold with probability at least $1 - K\delta$. The rightmost equality is obtained directly from the recursive update rule of $\hat{\A}_{\bar{t}_K(\delta)+1}$. Thus, we conclude:
        \[
            \prob\big(\hat\A_t = \{\theta^0\}\big) \ge 1 - K\delta, \quad \text{for } t = \tau(\delta) + 1.
        \]
        To show that this property holds for every $t \ge \tau(\delta) + 1$, we invoke Theorem~\ref*{thm:DAS-evolution}. Suppose there exists a time $t > \tau(\delta) + 1$ that is the first time after $\tau(\delta)+1$ at which ARL updates the DAS, i.e., the else-condition is triggered. Then, the updated DAS satisfies:
        \[
            \hat\A_t = \hat\A_{t-1} \cap \I(\hat{p}_t) = \hat\A_{\tau(\delta)+1} \cap \I(\hat{p}_{\tau(\delta)+1}) = \{\theta^0\}
        \]
        with probability at least $1 - K\delta$. The leftmost equality follows from Theorem~\ref*{thm:DAS-evolution}. The second equality follows because $t$ is the first update time after $\tau(\delta) + 1$, implying $\hat\A_{t-1} = \hat\A_{\tau(\delta)+1}$, as $t-1\le t$, and $\hat{p}_t = \hat{p}_{\tau(\delta)+1}$. The rightmost equality holds because $\hat{\A}_{\tau(\delta)+1} = \{\theta^0\}$ with probability at least $1 - K\delta$, and $\theta^0 \in \I(\hat{p}_{\tau(\delta)+1})$ by definition. Therefore, even if ARL updates the DAS after time $\tau(\delta)+1$, the DAS remains $\{\theta^0\}$ with probability $1 - K\delta$. This completes the proof of Part (2).
    \hfill$\Box$\endproof

\proof{\textbf{Proof of Theorem \ref*{thm:regret}.}} 
    Given $\delta \in (0,1]$, define the shorthand $\bar \tau := \min(\tau(\delta), T)$ and consider the following inequalities:
    \begin{align*}
        R_T(\piARL)
            & = \CIR(\piCI) - \CIR(\piARL) \nonumber \\  
            & = \expt^{\piCI} \left[ \sum_{t=1}^T \pRV_t^{\piCI} \cdot \sum_{i=1}^{N_t} \dRV_{ti}(\pRV_t^{\piCI}) \right] 
                -  \expt^{\piARL} \left[ \sum_{t=1}^T \pRV_t^{\piARL} \cdot \sum_{i=1}^{N_t} \dRV_{ti}(\pRV_t^{\piARL}) \right] \\
            & = \sum_{t=1}^T \left(\expt^{\piCI} \left[\sum_{i=1}^{N_t} \pRV_t^{\piCI} \cdot \dRV_{ti}(\pRV_t^{\piCI}) \right] 
                -  \expt^{\piARL} \left[ \sum_{i=1}^{N_t} \pRV_t^{\piARL} \cdot \dRV_{ti}(\pRV_t^{\piARL}) \right] \right) \\
            & = \sum_{t=1}^T N_t \left(\expt^{\piCI} \left[r(\pRV_t^{\piCI};\theta^0) \right] 
                -  \expt^{\piARL} \left[ r(\pRV_t^{\piARL};\theta^0)\right] \right) \\
            & = \sum_{t=1}^{\bar \tau} N_t \left(\expt^{\piCI} \left[r(\pRV_t^{\piCI};\theta^0) \right] 
                -  \expt^{\piARL} \left[ r(\pRV_t^{\piARL};\theta^0)\right] \right) \\
            & \hspace{2cm} + \sum_{t=\bar \tau+1}^T N_t \left(\expt^{\piCI} \left[r(\pRV_t^{\piCI};\theta^0) \right] 
                -  \expt^{\piARL} \left[ r(\pRV_t^{\piARL};\theta^0)\right] \right) \\
            & \le 2\bar{r} \sum_{t=1}^{\bar \tau} N_t.
    \end{align*}
    The first equality follows from the definition of regret $R_T(\piARL)$ in \S \ref*{subsec:ARL-theory}. The second equality applies the definition of $\CIR(\cdot)$ from \S \ref*{sec:Problem Formulation}. The third equality changes the order of expectation and summation, which is justified by Fubini’s theorem since all per-period expectations are finite under both policies. The fourth equality uses the fact that customers arrive i.i.d. within each period, so the expected revenue from $N_t$ customers at price $p$ is $N_t \cdot r(p; \theta^0)$. The fifth equality splits the summation over $T$ into two terms,  before and after $\bar{\tau}$, to isolate the contribution to regret prior to and following the demand identification time $\tau(\delta)$ (or $T$ if $\tau(\delta) = T+1$). The final inequality is obtained by bounding these two terms separately. The first term (which has the summation over $t\le\bar{\tau}$) is bounded above by $2\bar{r} \sum_{t=1}^{\bar \tau} N_t$ since for each $t$, the following inequalities hold:
    \begin{align*}
        & \expt^{\piCI} \left[r(\pRV_t^{\piCI};\theta^0) \right] 
                -  \expt^{\piARL} \left[ r(\pRV_t^{\piARL};\theta^0)\right]\\
        & \quad = \sum_{p\in\pSet} \left(\Pr(\pRV_t^{\piCI} = p) - \Pr(\pRV_t^{\piARL} = p)\right)r(p;\theta^0) \\
        & \quad \le 2\bar{r}
    \end{align*}
    In the above, the first equality follows directly from the definition of expectation for discrete random variables. The inequality then follows from the definition of $\bar{r}$ as the maximum absolute difference in expected revenues across all price pairs under $\theta^0$. For the second term (which involves the summation over $t \ge \bar{\tau}$), we invoke Proposition \ref*{prop:stoch-to-det}, which ensures that the DAS satisfies $\hat{\A}_{t} = \{\theta^0\}$ for all $t \ge \tau(\delta) + 1$ with probability at least $1 - K\delta$. From the second property of the risk-adjusted revenue function in Definition~\ref*{def:risk-adjust}, it follows that
    \[
        \eta(p; \hat\A_t) = r(p; \theta^0) \quad \text{whenever } \hat\A_t = \{\theta^0\}.
    \]
    Thus, for every $t \ge \tau(\delta) + 1$, the ARL price satisfies
    \[
        \hat{p}_t = \argmax\left\{\eta(p;\hat\A_t): p \in \PIP(\hat\A_t)\right\} = \argmax\left\{r(p; \theta^0): p \in \pSet\right\} = \pCI,
    \]
    where we used the identity $\PIP(\{\theta^0\}) = \pSet$ from Definition~\ref*{def:risk-adjust-rev}. Therefore, the ARL and CI policies coincide for all $t \ge \tau(\delta) + 1$, and the corresponding regret contribution is zero with probability at least $1 - K\delta$. Consequently, the regret for $t > \bar{\tau}$ is zero with probability at least $1 - K\delta$, yielding the bound.
\hfill$\Box$\endproof

\proof{\textbf{Proof of Corollary \ref*{cor:unconstrained}.}}
    Fix $\delta \in (0,1]$ and suppose $N_t = 1$ for every $t \in \indexSet{T}$, so that the total number of customer arrivals is $M = T$ and the firm can update prices after each arrival. From Theorem~\ref*{thm:regret}, with probability $1 - K\delta$, we know that the regret bound of the ARL policy is bounded above by
    \[
        R_T(\piARL) \ \le \ 2\bar{r} \cdot \sum_{t=1}^{\min\{\tau(\delta),T\}} N_t = 2\bar{r} \cdot \min\{\tau(\delta), T\},
    \]
    where we simplified the summation above using the fact that $N_t = 1$. The demand identification time $\tau(\delta)$ is given by $\bar{t}_K(\delta)$, where for each $k\le K$, we have:
    \[
        \bar{t}_k(\delta) := \min\bigg\{ \min\bigg\{t \in \{\bar{t}_{k-1}(\delta)+1, \dots, T\} : \sum_{s=\bar{t}_{k-1}(\delta)+1}^{t} N_s \ge n_\delta(\bar{p}_k) \bigg\},\, T+1 \bigg\}.
    \]
    Using the fact that $N_t=1$, we have $\bar{t}_1(\delta)= \min\big\{n_\delta(\bar{p}_1), T+1\big\}$,  $\bar{t}_2(\delta)= \min\big\{n_\delta(\bar{p}_1) + n_\delta(\bar{p}_2),  T+1\big\}$, and so on. Thus, for $k=K$, we obtain the following simplified formulation:
    \[
        \tau(\delta) = \bar{t}_K(\delta) = \min\left\{\sum_{k=1}^K n_\delta(\bar{p}_{k}),\ T+1\right\}.
    \]
    Replacing the above value of $\tau(\delta)$ in the regret bound of the ARL policy mentioned above, we obtain the following result that
    \[
        R_T(\piARL) \ \le \ 2\bar{r} \cdot \min\left\{\sum_{k=1}^K n_\delta(\bar{p}_{k}), \  T\right\},
    \]
    which holds with probability $1 - K\delta$. This completes the proof. 
\hfill$\Box$\endproof

\proof{\textbf{Proof of Proposition \ref*{prop:rev-ARL-NRM}.}}
    Fix $\delta \in (0,1]$, price $p \in \mathcal{P}$, and time $t \in \indexSet{T}$. Let $(\hat\A_{1}, \dots, \hat\A_{T})$ and $(\hat p_{1}, \dots, \hat p_{T})$ be arbitrary sequences of DAS and prices from Algorithm \ref*{alg:ARL}. Using Theorem \ref*{thm:DAS-evolution}, which guarantees that the inclusion $\theta_0 \in \hat\A_t$ holds with probability $1 - \delta$, we obtain the following inequalities:
    \begin{equation}\label{eq:ec:NRM-ARL-reve}
        r(p; \theta^0) \ = \ \eta(p;\{\theta^0\}) \ \ge \ \eta(p; \hat \A_{t}) \ \ge \ \eta(p; \Theta).
    \end{equation}
    The equality above follows from the second property in Definition \ref*{def:risk-adjust}. The leftmost inequality follows from the first property in this definition, noting that the inclusion $\{\theta^0\} \subseteq \hat\A_{t}$ holds with probability at least $1 - \delta$. The rightmost inequality follows from the same first property, applied to the inclusion $\hat\A_{t} \subseteq \Theta$, which holds with probability one. Therefore, with probability at least $1 - \delta$, the following inequality holds:
    \begin{align*}
        \big|\CIO(p; \theta^0) - \ARLO(p; \hat\A_t)\big| 
            & \ = \ |r (p; \theta^0)  - \eta(p; \hat\A_{t})|\\ 
            & \ = \   r (p; \theta^0)  - \eta(p; \hat\A_{t})\\
            & \ \le \ r (p; \theta^0) - \eta(p; \Theta)\\
            & \ = \  |r (p; \theta^0) - \eta(p; \Theta)|\\
            & \ = \   \big|\CIO(p; \theta^0) - \NRM(p; \Theta)\big|
    \end{align*}
    The above inequality follows directly from (\ref*{eq:ec:NRM-ARL-reve}) and the definitions of $\CIO(p; \theta^0)$, $\ARLO(p; \hat\A_t)$, and $\NRM(p; \Theta)$ in \S \ref*{sec:value-of-adaptation}. This inequality completes the proof because our choices of price $p \in \mathcal{P}$ and time $t \in \indexSet{T}$ are arbitrary. 
\hfill$\Box$\endproof

\begin{proposition}\label{prop:regret-NRM}
    The NRM policy's regret is upper bounded as follows:
    \[
        R_T(\piNRM) \ \le \ \bar{r}\cdot M\cdot \indicator\{\pCI \ne \pNRM\}.
    \]
\end{proposition}
\proof{\textbf{Proof of Proposition \ref*{prop:regret-NRM}.}} 
    Consider the following equalities:
    \begin{align*}
            R_T(\piNRM)
                & = \CIR(\piCI) - \CIR(\piNRM) \nonumber \\  
                & = \sum_{t=1}^T N_t \left(r(\pCI;\theta^0) - r(\pNRM;\theta^0)\right) \\
                & = M \left(\pCI\mu(\pCI;\theta^0) - \pNRM\mu(\pNRM;\theta^0)\right)\\
                & \le \bar{r}\cdot M\cdot \indicator\{\pCI \ne \pNRM\}
    \end{align*}
    The first equality follows from the definition of regret in \S \ref*{subsec:ARL-theory}. The second equality uses the fact that both the CI and NRM policies apply deterministic prices, $\pCI$ and $\pNRM$, and that customers arrive i.i.d. within each period, so the expected revenue from $N_t$ customers at price $p$ is $N_t \cdot r(p; \theta^0)$. The last inequality is obtained by considering two cases. First, if $\pCI = \pNRM$, then the CI and NRM revenues coincide, implying $R_T(\piNRM) = 0$, as reflected in the upper bound $\bar{r}\cdot M\cdot \indicator\{\pCI \ne \pNRM\}$. Second, if $\pCI \ne \pNRM$, then
    \[
            M \left(\pCI\mu(\pCI;\theta^0) - \pNRM\mu(\pNRM;\theta^0)\right)\le M \pCI\mu(\pCI;\theta^0)\le \bar{r}\cdot M.
    \]
    Thus, the upper bound holds in both cases, completing the proof.
\hfill$\Box$\endproof

\proof{\textbf{Proof of Proposition \ref*{prop:rev-risk-FTL}.}}% 
    Fix price $p\in\pSet$ and time $t\in\indexSet{T}$. The following inequalities hold:
    \[        
        \min_{\{\theta^0\}\subseteq\A \subseteq\Theta}\ARLO(p;\A) 
            = \min_{\{\theta^0\}\subseteq\A\subseteq\Theta}\eta(p; \A)
            \ \le \   \eta(p; \{\theta^0\}) 
            = r(p;\theta^0) = \CIO(p; \theta^0).
    \]
    The leftmost identity follows from the definition of $\ARLO(p;\A)$ in \S \ref*{sec:value-of-adaptation}. The inequality is a direct result of the first property in Definition \ref*{def:risk-adjust}, which ensures $\eta(p; \{\theta^0\}) \geq \eta(p; \A)$ since $\{\theta^0\} \subseteq \A$. The second equality from the left follows from the second property in Definition \ref*{def:risk-adjust}, which ensures $\eta(p; \{\theta^0\}) = r(p;\theta^0)$. The rightmost equality follows from the definition of $\theta^0$. Next, consider the following inequalities:
    \[
            \min_{\{\theta^0\} \subseteq \A \subseteq \Theta} \ARLO(p; \A) 
                \ = \ \eta(p; \A^*) 
                \ge \eta(p; \Theta)  
                \ge \min_{\theta \in \Theta} r(p;\theta).
    \]
    As above, the leftmost identity follows from the definition of $\ARLO(p;\A^*)$ in \S \ref*{sec:value-of-adaptation}, where $\A^*$ denotes the ambiguity set achieving the minimum in the optimization $\min\{\ARLO(p;\A): \{\theta^0\} \subseteq \A \subseteq \Theta\}$. The leftmost inequality follows from the first property in Definition \ref*{def:risk-adjust}, which ensures $\eta(p; \A^*) \geq \eta(p; \Theta)$ since $\A^* \subseteq \Theta$. The rightmost inequality holds because $\eta(p; \Theta)$ selects an element from the revenue set $\R(p; \Theta)$, while $\min_{\theta \in \Theta} r(p; \theta)$ is the minimum value in this set. Thus, either $\eta(p; \Theta) = \min_{\theta \in \Theta} r(p; \theta)$ or $\eta(p; \Theta) > \min_{\theta \in \Theta} r(p; \theta)$, depending on the specific form of the risk-adjusted revenue function.
\hfill$\Box$\endproof

\proof{\textbf{Proof of Proposition \ref*{prop:FTL-param}.}}
    We observe that both ARL and FTL, outlined in Algorithms \ref*{alg:ARL} and \ref*{alg:FTL}, respectively, update the DAS $\hat\A_t$ and point estimate $\thetaFTL{t}$ at the same iterations. That is, if the if-condition (or else-condition) in ARL is triggered at time $t$, the same occurs in FTL. From the proof of Theorem \ref*{thm:DAS-evolution}, recall that with probability at least $1 - \delta$, we have $\ell\big(\theta ; \D(\hat p_t)\big) \le c(\hat p_t)/2$ for every $\theta \in \I(\hat p_t)$ and $\ell\big(\theta ; \D(\hat p_t)\big) > c(\hat p_t)/2$ for every $\theta \in \Theta \backslash \I(\hat p_t)$. This property can be directly tailored to FTL by replacing the ARL price $\hat p_t$ with the FTL price $\pFTL_t$, yielding the following inequality:
    \begin{equation}\label{eq:ec:FTL-ineq}
        \prob\left( \max_{\theta \in \I(\pFTL_t)} \ell(\theta; \D(\pFTL_t)) \le \frac{c(\pFTL_t)}{2}, \ \max_{\theta \in \Theta \backslash \I(\pFTL_t)} \ell(\theta; \D(\pFTL_t)) > \frac{c(\pFTL_t)}{2} \right) \ge 1 - \delta.
    \end{equation}
    Now, if at time $t \in \indexSet{T}$ the else-condition in FTL occurs, then FTL updates its estimate $\thetaFTL{t}$ to 
    \[
        \thetaFTL{t+1} = \argmin\big\{\ell\big(\theta; \D(\pFTL_t)\big) : \theta \in \Theta\big\}.
    \]
    Combining (\ref*{eq:ec:FTL-ineq}) with this definition of $\thetaFTL{t+1}$, which is the parameter in $\Theta$ that minimizes the loss $\ell\big(\theta; \D(\pFTL_t)\big)$, we conclude that $\thetaFTL{t+1}$ cannot lie in $\Theta \backslash \I(\pFTL_t)$, as models in this set have higher loss than those in $\I(\pFTL_t)$ with high probability. Hence, with probability $1 - \delta$, we have $\thetaFTL{t+1} \in \I(\pFTL_t)$.
\hfill$\Box$\endproof

\proof{\textbf{Proof of Theorem \ref*{thm:regret-FTL}.}}
    \textbf{\underline{Part (1).}}% 
        We show that the identity $\hat{\A}_t = \{\thetaFTL{t}\} = \{\theta^0\}$ holds with probability $1 - \delta$. Proposition \ref*{prop:stoch-to-det} establishes that for every $t \in \{\tau(\delta)+1, \dots, T\}$, we have $\hat{\A}_t = \{\theta^0\}$ with probability at least $1 - K\delta$. Under Assumption \ref*{asm:separability}, which is required in Proposition \ref*{thm:regret-FTL}, we have $K = 1$. Hence, with probability $1 - \delta$, it holds that $\hat{\A}_t = \{\theta^0\}$ for all $t \ge \tau(\delta) + 1$. It remains to show that $\thetaFTL{t} = \theta^0$ holds with the same probability. Consider time $t = \tau(\delta)$. From the definition of $\tau(\delta)$ under Assumption \ref*{asm:separability}, we have
        \[
            \tau(\delta) = \bar{t}_1(\delta) = \min\bigg\{ \min\bigg\{t \in \{1, \dots, T\} : \sum_{s=1}^{t} N_s \ge 4 \cdot \max\left\{2\left(\frac{v}{\hat{c}}\right)^2, \frac{b}{\hat{c}}\right\} \cdot \ln\left(\frac{2}{\delta}\right) \bigg\}, \ \ T+1 \bigg\}.
        \]
        This definition shows that $\tau(\delta)$ is the first time the minimum-data condition $\sum_{s=1}^{t} N_s\ge n_\delta(\pFTL_1)$ is met. That is, from $t = 1$ to $t = \tau(\delta)$, FTL plays the fixed price $\pFTL_1 = \argmax\{r(p;\thetaFTL{1}) : p \in \pSet\}$. At time $t = \tau(\delta)$, the else-condition in FTL is triggered, and FTL updates its estimate. By Proposition \ref*{prop:FTL-param}, we then have the probabilistic inclusion $\thetaFTL{t+1} \in \I(\pFTL_1) = \{\theta^0\}$, so
        \[
            \prob\left(\thetaFTL{t} = \theta^0\right) \ge 1 - \delta, \quad \text{for } t = \tau(\delta)+1.
        \]
        From time $t = \tau(\delta)+1$ onward, FTL does not update $\thetaFTL{t}$ until the minimum-data condition is met again. If an update occurs at a future time $t > \tau(\delta)+1$, Proposition \ref*{prop:FTL-param} again ensures $\thetaFTL{t} = \theta^0$ with probability at least $1 - \delta$. Therefore, for all $t \ge \tau(\delta)+1$, we conclude:
        \[
            \prob\left(\hat{\A}_t = \{\thetaFTL{t}\} = \{\theta^0\} \right) \ge 1 - \delta.
        \]
    \textbf{\underline{Part (2).}} %
        We show that $R_T(\piFTL) \le 2\bar{r} \sum_{t=1}^{\min(\tau(\delta), T)} N_t$. Intuitively, since $\thetaFTL{t} = \theta^0$ holds with probability $1 - \delta$ for every $t \ge \tau(\delta)+1$, the FTL policy's regret is upper bounded by the data collected from $t = 1$ to $t = \min(\tau(\delta), T)$. The formal proof follows the same steps as the proof of Theorem \ref*{thm:regret}, with adjustments to replace $\piARL$ with $\piFTL$ throughout. We do not repeat the proof here accordingly.
\hfill$\Box$\endproof

\end{document}